%% file: main.tex
\documentclass[letterpaper,11pt]{article}

\usepackage[english]{babel}
\usepackage[utf8]{inputenc}
\usepackage{amsmath}
\usepackage{graphicx}
\date{}
\title{
\vspace{-41pt}
\vspace{-0.5cm}Select without Fear: \\Almost All Mini-Batch Schedules Generalize Optimally
}
\author{\vspace{-2cm}}
        
\usepackage[T1]{fontenc}
\usepackage{lmodern}
\usepackage{cite}
\usepackage{appendix}

\PassOptionsToPackage{hyphens}{url}\usepackage{hyperref}%
\hypersetup{
    colorlinks=true,
    linkcolor=black,
    filecolor=magenta,      
    urlcolor=cyan,
    citecolor=blue	
}

\usepackage{colortbl}
\usepackage{tabulary}
\usepackage{etoolbox}
\usepackage[hyphens]{url}
\usepackage[hyphenbreaks]{breakurl} 

\usepackage{booktabs}       
\usepackage{amsfonts}       
\usepackage{nicefrac}       
\usepackage{microtype}      
\usepackage{verbatim}
\usepackage{amsmath}
\usepackage{amssymb}
\usepackage{stackrel}
\usepackage{verbatim}
\usepackage{url}
\usepackage{algpseudocode}
\usepackage{graphicx}
\usepackage{enumerate}
\usepackage{algorithm}
\usepackage{algpseudocode}
\usepackage{algorithm,algpseudocode}
\let\oldReturn\Return
\renewcommand{\Return}{\State\oldReturn}
\usepackage{float}
\usepackage{blindtext}
\usepackage{multirow}
\usepackage{dsfont}
\usepackage{booktabs}
\usepackage{mathtools}
\newtheorem{assumption}{Assumption}
\newcommand{\qedwhite}{\hfill \ensuremath{\Box}}

\usepackage[symbol]{footmisc}

\usepackage[table]{xcolor}
\input{macros}
\usepackage{enumitem}
\usepackage{amssymb}
\usepackage{pifont}

\usepackage{tikz}
\setitemize{noitemsep}
 \usepackage{makecell}

\usepackage{mathtools}
\DeclarePairedDelimiterX{\inp}[2]{\langle}{\rangle}{#1, #2}
\usepackage{geometry}
\allowdisplaybreaks
\newcolumntype{P}[1]{>{\centering\arraybackslash}p{#1}}
\newcolumntype{M}[1]{>{\centering\arraybackslash}m{#1}}
\newcommand{\BlackBox}{\rule{1.5ex}{1.5ex}}  

\newtheorem{theorem}{Theorem}
\newtheorem{lemma}[theorem]{Lemma} 
 
\newtheorem{remark}[theorem]{Remark}

\newtheorem{definition}[theorem]{Definition}

\usepackage{nicematrix}

\begin{document}
\maketitle
\begin{center}
\vspace{-1.5cm}
\begin{tabular}
{ >{\centering}m{5cm}  >{\centering}m{5cm}}
\multicolumn{2}{c}{Konstantinos E. Nikolakakis}\\
\multicolumn{2}{c}{\texttt{\small konstantinos.nikolakakis@yale.edu}} 
\tabularnewline
\multicolumn{2}{c}{\vspace{-6pt}}
\tabularnewline
Amin Karbasi \\ 
\texttt{\small amin.karbasi@yale.edu} 
& 
Dionysis Kalogerias \\
\texttt{\small dionysis.kalogerias@yale.edu}
\tabularnewline
\end{tabular}
\vspace{2.5bp}
\end{center}


\begin{abstract}
 We establish matching upper and lower generalization error bounds for mini-batch Gradient Descent (GD) training with either deterministic or stochastic, data-independent, but otherwise arbitrary batch selection rules. We consider smooth Lipschitz-convex/nonconvex/strongly-convex loss functions, and show that classical upper bounds for Stochastic GD (SGD) also hold verbatim for such arbitrary nonadaptive batch schedules, including all deterministic ones. Further, for convex and strongly-convex losses we prove matching lower bounds directly on the generalization error uniform over the aforementioned class of batch schedules,
 showing that all such batch schedules generalize optimally. Lastly, for smooth (non-Lipschitz) nonconvex losses, we show that full-batch (deterministic) GD is essentially optimal, among all possible batch schedules within the considered class, including all stochastic ones. 
\end{abstract}
\vspace{-0.1cm}\textbf{Keywords:} Generalization Error, Minimax Bounds, Smooth Nonconvex/Convex Optimization
\section{Introduction}
Stochastic gradient descent (SGD) constitutes one of the pillars of optimization theory and practice, receiving widespread attention for more than $60$ years. There is a long line of work focusing on optimization error and convergence analysis of gradient-based algorithms. However, despite the established success of GD and its variants in optimization, the generalization abilities of gradient-based training schemes in the context of machine learning are still not quite well-understood. The celebrated work of Hardt et al. \cite{hardt2016train} on final iterate generalization bounds for standard SGD shed some light on this issue by focusing on a canonical class of learning problems with Lipschitz and smooth losses. Further, \cite{hardt2016train} also laid the foundation for exploring generalization error guarantees for other variants of SGD including SGD with early momentum~\cite{ramezani2021generalization}, randomized coordinate descent~\cite{wang2021stability}, look-ahead approaches~\cite{zhou2021towards}, noise injection methods~\cite{xing2021algorithmic}, and Stochastic Gradient with Langevin Dynamics (SGLD)~\cite{pensia2018generalization,mou2018generalization,li2019generalization,NEURIPS2019_05ae14d7,farghly2021time,wang2021optimizing,wang2021analyzing}. In parallel, a large number of studies on algorithmic stability~\cite{fu2023sharper,NEURIPS2022_fb8fe6b7,a240701fc9f84943820a45e3f36de2f8,kim2021black,pmlr-v80-kuzborskij18a,pmlr-v119-lei20c,Zhou2022,Li2020On,lei2020sharper,feldman2019high,madden2020high,klochkov2021stability,NIPS2016_8c01a759,bassily2020stability,lei2021generalizationMP,zhang2021stability,NEURIPS2022_cce0df2e} emerged through the special connection between uniform stability~\cite{NEURIPS2018_05a62416,pmlr-v125-bousquet20b} and generalization~\cite{bousquet2002stability,hardt2016train}. 

In this work, we look at the generalization error of general mini-batch GD training schemes (with no momentum and) with arbitrary data-independent batch schedules (see Algorithm \ref{alg:gen_SGD} and Definition \ref{eq:set_of_algos} below for details). In a nutshell, we extend classical upper bounds on the generalization error of SGD~\cite{hardt2016train} to all such general gradient based schemes, and derive matching, fully parameterized lower bounds directly on the generalization error, showing tightness of bounds and optimality of methods. A detailed statement of our contributions, which are also summarized in Table \ref{table:results_app}, is as follows:
\begin{itemize}
    \item  For smooth Lipschitz-convex/nonconvex/strongly-convex learning problems, we establish sharp upper generalization error bounds for a large class of gradient-based algorithms, namely those encompassing arbitrary stochastic or deterministic, data-independent selection rules of any batch size (including arbitrarily time varying batch schedules). 
    In the case of Lipschitz (resp. path-gradient-stable) and smooth (resp. strongly-)convex losses, our upper bounds hold essentially verbatim to those in prior work \cite{hardt2016train}, initially developed for SGD with batch size $1$. Further, in constrast  with \cite{hardt2016train}, we avoid a bounded loss assumption in the nonconvex case.

    \item We establish minimax-optimal matching corresponding lower bounds that hold uniformly over the whole class of algorithms under consideration (Definition \ref{eq:set_of_algos}) 
    and depend explicitly on the parameters of the learning problem under consideration (for instance Lipschitz and smoothness constants), the training horizon and the step size. Our lower bounds confirm the sharpness of the corresponding upper bounds, and characterize the \textit{optimal rate of algorithmic generalization} for each of the learning problem classes under study.\begin{table}[bt!]\label{table:results_app}
\centering
\begin{tabular}{|>{\centering}m{3.2cm}||>{\centering}m{4.3cm}||>{\centering}m{3.65cm}||>{\centering}m{2.9cm}|}
\hline 
\multicolumn{4}{|c|}{Generalization Rates for Gradient-Based Training --- valid for all $A_\mc{R}\in\setalgos$ ---}\tabularnewline
\hline 
\vspace{4pt}
Stepsize $\eta_t$
\vspace{4pt}
& 
\vspace{4pt} 
Upper Bound on 
\\\!$\stackrel[(f,\mc{D})\in \mc{C}]{}{\sup}\!|\gen (f,\mc{D},A_{\mc{R}}) |$
\vspace{4pt} 
& 
\vspace{6pt}
Lower Bound on
\\\!$\stackrel[(f,\mc{D})\in \mc{C}]{}{\sup}\!|\gen (f,\mc{D},A_{\mc{R}}) |$
\vspace{2pt}
&\vspace{4pt}
Problem Class $\mc{C}$
\vspace{4pt}
\tabularnewline
\hline
\specialrule{1.5pt}{0.5pt}{0.5pt}
\hline
$\eta_t \le 2/\beta$ & 
\cellcolor{blue!6} 
\vspace{3bp} $\displaystyle{\frac{2L^2}{n}  \sum^{T}_{t=1}  \eta_t }$ 
\vspace{-0bp}
& 
\cellcolor{blue!6}
\vspace{3bp} $\displaystyle{\frac{L^2}{ 2n} \sum^{T}_{t=1} \eta_t}$
\vspace{-0bp}
& Convex \\ $L$-Lipschitz \\ $\beta$-Smooth  
\tabularnewline 
\hline 
$\eta_t\leq C/ t $, $ C <  1/\beta$ &
\cellcolor{blue!6}\vspace{-10bp}
$\dfrac{2 C( 1+ \frac{1}{C \beta}) e^{C\beta} L^2 T^{C\beta }}{n} $ \vspace{-10bp} & \textbf{OPEN} &
Nonconvex \\ $L$-Lipschitz \\ $\beta$-Smooth
\tabularnewline 
\hline 
$\eta_t\leq c/ \beta t$,   
$c \le 1 $
& 
$\hspace{-4.5pt}\mc{O}\hspace{-2.5pt} \lp \dfrac{ T^c\sqrt{\log (T)}}{n} \rp$ \hspace{-7pt} (\hspace{-0.5pt}\cite{nikolakakis2023beyond},\hspace{-2pt} GD)\hspace{-5pt}\vspace{1pt} 
&
\cellcolor{blue!6}
\vspace{-22pt}
$ \dfrac{(T+1)^{\log(1+c)} -1    }{2n  }   
    $
%
&
\vspace{6pt}
Nonconvex \\ $\beta$-Smooth
\vspace{4pt}
\tabularnewline
\hline
$\eta_t =\eta\in [\ell_1 ,\ell_2 ]$ \\ \vspace{+0.1cm} 
$\ell_1 \triangleq  2/\gamma(T+1)$\\
$\ell_2 \triangleq 1/(\beta+\gamma)$
& \cellcolor{blue!6} 
\vspace{-14bp}
$\dfrac{4 {L}^2}{n \gamma } $ \vspace{-10bp} & \cellcolor{blue!6} \vspace{-0bp}
$\hspace{6pt} \dfrac{L^2}{32\gamma n}  $ \vspace{-3.5bp}& 
\!\!$\gamma$-strongly-convex\\$\beta$-Smooth\\$L$-\textit{path}-Lipschitz  
\tabularnewline
\hline 
\end{tabular}
\caption{For certain stepsize choices, uniform over the algorithm class $\setalgos$ (see Definition \ref{eq:set_of_algos}) upper and lower bounds on the worst-case generalization error $\sup_{(f,\mc{D})\in \mc{C} } |\gen (f,\mc{D},A_{\mc{R}}) |$ appear in the second and third columns, respectively, where the class of learning problems $\mc{C}$  under consideration appears in the last column; here, $f$ and $\mc{D}$ denote the loss function and data distribution, respectively, and $A_{\mc{R}} \in \setalgos$. All colored boxes indicate our contributions, whereas the upper bound for GD on nonconvex smooth losses is established in prior work~\cite{nikolakakis2023beyond}. Establishing a lower bound for the class of Lisphitz and smooth nonconvex losses remains an open problem left for future work.
}
\end{table}

    \item For Lipschitz (resp. path-gradient-stable) and smooth (resp. strongly-)convex losses, we show that all mini-batch GD methods with data-independent batch schedules generalize optimally. This fact in particular rigorously implies \textit{no competitive advantage of stochastic training over deterministic training}, for the aforementioned class of learning problems. In other words, deterministic training provably performs as optimally as stochastic training in terms of generalization performance. 

    \item In the same spirit, we show that \textit{full-batch (deterministic) GD} is optimal within the class of smooth (and possibly non-Lipschitz nonconvex) losses. In fact, our lower bound essentially matches the upper bound from our prior work in \cite{nikolakakis2023beyond}. While optimality of full-batch GD training within the class of smooth (possibly non-Lipschitz, nonconvex) losses is established in this work, whether stochastic training is also optimal in the setting remains an open problem.

\end{itemize}


\color{black}

\subsection{Comparison with Prior Works}
First, we explain how our analysis provides optimal generalization error rates for general data-independent batch schedules, while standard techniques from prior works fail in this respect, namely the main technical approach taken in the seminal work of Hardt et al. in \cite{hardt2016train}. Then we compare results from prior works with ours, also commenting on the now provable fact that stochastic training has essentially no competitive advantage against deterministic training, under very general assumptions. Finally, we provide a detailed discussion regarding the lower bounds developed in this work, and how they are positioned within the related literature. 

\subsubsection{Suboptimality of Uniform Stability} 

We begin by demonstrating that the uniform stability approach as well as the core proof technique devised in~\cite{hardt2016train} (as expected) result in \textit{vacuous} bounds for a non-negligible set of nontrivial training algorithms. At the same time, the on-average stability approach advocated in this work produces optimal (and thus correct) bounds, \textit{uniformly} within the postulated class of algorithms. Even though both uniform and on-average stability result in tight bounds for vanilla SGD~\cite{pmlr-v119-lei20c}, uniform stability ``throws away too much structure to start with'' and is thus inadequate for analyzing other common batch-schedules and variants of (S)GD. To illustrate this further with a simple argument, we prove here that the uniform stability approach of ~\cite{hardt2016train} indeed fails for the deterministic incremental (round-robin) gradient method \cite[Algorithm 3]{pmlr-v125-safran20a},~\cite{bertsekas1999nonlinear,bertsekas2015convex,Nedić2001,doi:10.1137/17M1147846,doi:10.1137/15M1049695}, which among various other applications has been used to train neural networks for more than $40$ years. Later, we  demonstrate that the proof technique utilized in~\cite{hardt2016train} also gives vacuous generalization bounds for general Lipschitz and smooth convex and strongly-convex losses (see Appendix \ref{failure}).


Specifically, consider the  loss $f(w,z) = \sum^{d}_{k=1} w^k z^k$, with $w\in\mbb{R}^d$ and ($d$ dimensional) examples $z_i\triangleq (z^1_i, z^2_i ,\ldots ,z^{d}_i )\in \{-1,+1 \}^d, i\in\{1,\ldots,n\}$. Then, for the incremental gradient method (see, e.g., ~\cite[Algorithm 3]{pmlr-v125-safran20a}), denoted here as a map $A(\cdot)$ on example sequences, and for neighboring example sequences $S,S'$~\cite{hardt2016train}, the uniform stability constant of \cite[Theorem 2.2]{hardt2016train} may be calculated explicitly (while assuming a decreasing step size sequence which restarts after each epoch to the initial value $\eta_1$) as
\begin{align*}
    \sup_{S, S',z} \E_{A} [f(A(S);z)-f(A(S');z)] &= \sup_{S, S',z}  [f(A(S);z)-f(A(S');z)]\numberthis \label{eq:fck_BMDY}\\
    &= \sup_{S, S',z} \sum^{d}_{k=1} (A^k  (S) - A^k  (S') ) z^k  \\
    & = \sup_{S, S',z}\sum^{d}_{k=1} \lp w^k_1- \sum^{T}_{t=1}\eta_t\sum_{\mc{Z}\in J_t} \mc{Z}^k -  w^k_1+ \sum^{T}_{t=1}\eta_t\sum_{\mc{Z}\in J'_t} \mc{Z}^k \rp z^k \numberthis \label{eq:f1}
     \\
    & = \sup_{S, S',z} \sum^{d}_{k=1} \sum^{T}_{t=1}\eta_t\lp (z_t')^k - z^k_t  \rp z^k \numberthis \label{eq:f2}\\
    & = K\times \sup_{S, S', z} \sum^{d}_{k=1} \eta_1\lp (z_1')^k - z^k_1  \rp z^k = 2K d \eta_1,\numberthis \label{eq:f3}
\end{align*} where $J_t, J'_t$ are the batches at time $t$ from the sequences $S,S'$ respectively. Equality \eqref{eq:f2} comes from the size of the batch $|J_t|=|J'_t|=1$. Lastly, equality \eqref{eq:f3} holds for any $K\in\mbb{N}$ number of epochs, and the supremum is achieved for the sequences sequences $S,S'$ that differ in the first entry because the step-size sequence is assumed non-decreasing (at each epoch). Evidently, because the uniform stability is a constant independent of the number of samples $n$, this approach fails completely to provide a meaningful generalization bound, even in this simple example. In fact, the \textit{optimal} generalization bound for the incremental gradient method in this case is $\frac{2 d}{n} \sum^T_{t=1}\eta_t $, for every choice of training horizon $T$ and stepsize (sequence) $\eta_t$. We refer the reader to Appendix \ref{failure} for a more detailed and general comparison.
It is also easy to see that the uniform stability approach fails more generally within the general class of Lipschitz and smooth losses by directly evaluating \eqref{eq:fck_BMDY} (also see Appendix \ref{failure}).

Overall, it becomes plain that uniform stability constants in general \textit{cannot be} sufficient to characterize algorithmic generalization (except for certain cases such as standard SGD studied in \cite{hardt2016train}).
Rather than utilizing uniform stability, in this work we show that the (or a) correct technical approach is that of bounding the generalization error in terms of \textit{on-average algorithmic stability}, providing tight bounds which characterize algorithm generalization in a \textit{minimax-optimal sense} (see Section \ref{minimax_sec}).

\subsubsection{Stochastic Training is not Necessary}
 Many prior works consider stochastic batch scheduling and some even suggest that stochasticity is necessary for good generalization~\cite{hardt2016train,charles2018stability}. However more recent experimental~\cite{geiping2021stochastic} and theoretical~\cite{nikolakakis2023beyond} works suggest that deterministic training (e.g. full-batch GD) may be at least as good as stochastic training. Our findings imply that, at least within the learning settings considered, randomization in training provably offers no competitive advantage as compared with deterministic training in terms of the achievable generalization error (of course, the situation is different as far as the corresponding optimization error is concerned~\cite{pmlr-v98-wang19a}). An important implication of our results is that different mini-batch schedules mainly affect the optimization error (as appears in~\cite{pmlr-v98-wang19a}) rather than the generalization error for the class of Lipschitz and smooth losses. We also note that in the smooth (possibly non-Lipschitz, nonconvex) case, the question of optimality of batch schedules other than that of full-batch GD (which we prove to be optimal) remains open.

From a technical standpoint, our generalization error analysis in this paper is developed in a fully unified manner relative to the choice of the particular batch schedule. More specifically, we establish bounds which are uniform over the class of mini-batch GD training schemes with either deterministic or stochastic, data-independent, but otherwise arbitrary batch schedules. This class of batch schedules includes both deterministic and stochastic training as special cases. For instance, the set of algorithms supported by our analysis includes but is not limited to classical SGD with mini-batch of any size $m$, full-batch GD, round-robin deterministic batch selection, SGD with random reshuffling~\cite{Gürbüzbalaban2021}, SGD with single shuffling, and incremental gradient methods~\cite{pmlr-v125-safran20a,pmlr-v89-nacson19a,pmlr-v98-wang19a,mohtashami2022characterizing,doi:10.1137/15M1049695}.

\subsubsection{Lower Generalization Error Bounds (over Data-Independent Batch Schedules)}
To evaluate the performance of the algorithms under consideration, we are interested in deriving minimax ---over mini-batch GD algorithms ($\inf$) and learning problem instances ($\sup$)--- lower bounds \textit{directly} on the generalization error. Prior works have already developed lower bounds on uniform stability constants (while taking a supremum over the data-set) for non-smooth losses~\cite{bassily2020stability} and Lipschitz-smooth losses~\cite{zhang2022stability}, for SGD and full-batch GD. However, lower bounds on uniform stability do not imply a lower bound on the generalization error; as also demonstrated above, uniform stability is sufficient but not necessary for generalization. Further, recent work on the class of smooth convex losses~\cite{zhang2023lower} considered the construction of specific --and not fully parameterized-- problem instances to derive lower bounds on the corresponding excess risk (i.e., generalization $+$ optimization), and --in this sense-- showed tightness for the cases of SGD~\cite{pmlr-v119-lei20c,pmlr-v178-schliserman22a} and full-batch GD~\cite{nikolakakis2023beyond}. The construction of an elaborate example to lower bound the excess risk of the (S)GD algorithm also appeared in~\cite{amir2021sgd,amir2021never}. However, that construction consists of a single instance of a Lipschitz non-smooth convex loss with dimension $d \gtrsim 2^n$, where $n$ is the data-set size; this example is not only restrictive but also not reasonable because the number of dimensions is exponential in comparison with the size of data-set. On the other hand, for Lipschitz possibly non-smooth generalized linear models, \cite{NEURIPS2022_9521b6e7} recently shows that GD is optimal.

In contrast with prior works, we develop lower bounds over classes of learning problems parameterized by Lipschitzness and smoothness constants, as well as data-set size, step-size and training horizon. Our lower bounds hold over these parameterized classes of learning problems, uniformly over stochastic and deterministic gradient-based algorithms.
Our lower bounds essentially match our upper bounds, as can be seen in the third column of Table \ref{table:results_app}, for every choice of the set of related parameters (up to constants or inconsequential factors, e.g., $\sqrt{\log}$). As an additional benefit of our approach,
while information theoretic generalization error bounds are reportedly not tight for final-iterate algorithmic generalization~\cite{livni2023information,pmlr-v201-haghifam23a,xu2017information}, here we confirm that stability analysis~\cite{bousquet2002stability,hardt2016train} is provably tight for large classes of loss functions and gradient-based algorithms, very relevant for practical use. At this point, it is worth mentioning that our results (optimal upper and lower bounds) also hold for the popular (S)GLD (implemented using any data-independent batch schedule), which has been widely studied in the context of information theoretic algorithmic generalization.  

Consequently, (on-average) stability seems most probably one right approach to study algorithmic generalization. As also stated above, through our minimax lower bounds we prove the simultaneous optimality (or equivalence) of all mini-batch GD schemes with data-independent batch schedules for Lipschitz, smooth and convex, and for smooth and strongly-convex losses. For learning problems involving smooth and possibly nonconvex losses, we show that the (deterministic) full-batch GD algorithm is essentially optimal, among all possible gradient-based schemes considered (with possibly randomized batch schedules).  
%
%
%
%
%
%
%
%
%
\subsection{Preliminaries}\label{sec:preli}
Let $z_1,z_2,\ldots,z_n , z'_1,z'_2,\ldots,z'_n$ be i.i.d random variables, with respect to an unknown distribution $\mc{D}$. For brevity, define the sets $S\triangleq (z_1,z_2,\ldots,z_n)$ and $S^{(i)}\triangleq (z_1,z_2,\ldots,z'_i,\ldots,z_n)$ that differ at the $i^{\mathrm{th}}$ random element, and let us denote $S'\triangleq (z'_1,z'_2,\ldots,z'_n) $. Let $f(w,z)$ be the loss at the point $w\in\mbb{R}^d$ for some example $z\in\mc{Z}$. Given a data-set $ \{z_i\}^n_{i=1}$, our goal is to find the parameters $w^*$ of a learning model such that $ w^* \in \arg \min_{w} R(w)$, where $R(w)\triangleq \E_{Z\sim \mc{D}} [ f(w,Z) ]$ and $R^*\triangleq  R(w^* ) $.
%
%
Since the distribution $\mc{D}$ is not known, we consider the empirical risk 
\begin{align}
    R_S (w) \triangleq \frac{1}{n} \sum^n_{i=1} f(w ; z_i).
\end{align} 
The corresponding empirical risk minimization (ERM) problem is to find $w^*_S \in \arg \min_{w} R_S (w)$ (assuming minimizers on data exist for simplicity). 
 For a (stochastic) algorithm $A$ with input $S$, and output $A(S)$, the generalization error $\epsilon_{\mathrm{gen}}$ is defined as the difference between the empirical and population loss 
\begin{align}
   \gen &\triangleq \E_{S,A} [R(A(S)) -R_S (A (S)) ] = \E_{S,A,z'_i} \Big[  f(A (S); z'_i ) -  \frac{1}{n}\sum^n_{i=1} f(A (S); z_i ) \Big]. \label{eq"gen_def}
\end{align} We continue by introducing the class of gradient-based algorithms that we consider in this work. 
\begin{algorithm}
    \begin{algorithmic}[1]
    \caption{$A_{\mc{R}}$: Mini-batch (S)GD with Generic Batch Schedules}
	\label{alg:gen_SGD}
	\Require data-set $S= (z_1,z_2,\ldots, z_n)\in\mc{Z}^n$, horizon $T$, batch-size $m$, stepsize $\{ \eta_t\}^T_{t=1}$, loss function $f:\mbb{R}^d \times \mc{Z} \rightarrow \mbb{R}^+$.
	\State Choose initial point $w_1\in \mbb{R}^d$, possibly random but independent of $\cal D$. 
    \State \textbf{for} $t=1$ to $T$ \textbf{do}
        \State \hspace{+0.4cm} Choose $m\leq n$ distinct indices $\{k^t_1,k^t_2,\ldots,k^t_m\}\subseteq [n]$ according to a rule $\mc{R}$ (at $t$),
        
        $\hspace{-6pt}$possibly randomized but independent of $\mc{D}$.
        \State \hspace{+0.4cm} Set mini-batch $J_t\leftarrow \{z_{k^t_1}, z_{k^t_2},\ldots , z_{k^t_m} \}$. 
        \State \hspace{+0.4cm} $w_{t+1}  = w_t -  \frac{\eta_t}{m}\sum_{z\in J_t} \nabla f (w_t , z)$. 
    \Return $W_{T+1}$
      \end{algorithmic}
\end{algorithm}
\subsection{(Stochastic) Gradient Descent with Arbitrary Batch Schedules}
We study a wide class of mini-batch GD training algorithms, for any mini-batch selection rule (deterministic or stochastic) that is independent of the input data-set. For instance, this class includes the classical SGD with mini-batch of any size $m$, full-batch GD, deterministic approaches of sequential batch selection rules as well as randomized selection rules with arbitrary distribution for the batch choice (not necessarily identically distributed choices with respect to each sample), round-robin with deterministic selection, SGD with random reshuffling, SGD with single shuffling, or incremental gradient methods~\cite{pmlr-v125-safran20a}. In all cases, the mini-batch selection policy at time $t$ is independent of input data-set. The size of the mini-batch is $m$ and is considered to be fixed. Our analysis can be extended to time dependent batch selection size $m_t$ without affecting the analysis or the results, however, we consider it fixed for simplicity. 
A formal definition of the class of algorithms considered is as follows (see also the Algorithm \ref{alg:gen_SGD}).
\begin{definition}[Class of Algorithms]\label{eq:set_of_algos}
Given the number of iterations $T$ and a step-step size sequence $\{\eta_t\}^T_{t=1}$, we define the set of algorithms given by Algorithm \ref{alg:gen_SGD} for all batch-selection rules $\mc{R}$ (independent of the input data-set) as\footnote{We implicitly assume an underlying probability triplet on a sample space $\Omega$.} \begin{align}
    \setalgos\triangleq \{A_{\mc{R}} \normalfont\textrm{ as in Algorithm \ref{alg:gen_SGD}}\, | \,\mc{R}:\Omega \rightarrow  \{1,2,\ldots,n\}^{m\times T},\, 
    \P_{(\mc{R},S)}=\P_{\mc{R}}\P_S \}.
\end{align} 
\end{definition} In other words, the class of algorithms $\setalgos$ contains stochastic and deterministic batch selection rules such that each example may appear at most once in the batch $J_t$ with any probability (invariant of the data-set). For the rest of the paper we write $A_{\mc{R}}$ 
to denote an algorithm in the set $\setalgos$.
\subsection{On-Average Stability}
 A key part of our analysis is the derivation of upper bounds by considering on-average algorithmic stability, namely $\frac{1}{n}\sum^n_{i=1} \Vert  A_{\mc{R}} (S) - A_{\mc{R}} (S^{(i)})  \Vert$. Through on-average stability we show upper bounds for the generalization error by using a unified analysis among all algorithms $A_\mc{R}$ within $\setalgos$. Specifically, we develop a uniform stability analysis for all data-sets $S$ and selection rules $\mc{R}$. In contrast to prior work~\cite{hardt2016train} that considers only uniformity with respect to the data-set, we extend the generalization error upper bounds for a general class of gradient-based algorithms ($\setalgos$).

In line with standard prior work, for convex and nonconvex losses we assume that the loss is uniformly Lipschitz. For strongly-convex losses, the Lipschitz assumption does not hold and we instead consider a relaxed \textit{path-boundedness} assumption on the loss gradients, introduced later to avoid clutter (see Section \ref{sclsection}).
\begin{assumption}[Lipschitz Loss]\label{ass:Lip} There exists a constant $L\geq 0$, such that for all $w,u\in\mbb{R}^d$ and $z\in \mc{Z}$, $\Vert  f( w, z) -  f( u, z) \Vert_2 \leq L \Vert w- u \Vert_2 $.
\end{assumption} Through Assumption \ref{ass:Lip}, equation \eqref{eq"gen_def} gives an upper bound on the generalization error as follows \begin{align*}
   |\gen | &\triangleq \Big| \E_{S,z'_i,\mc{R}} [ f(A_{\mc{R}} (S); z'_i )] - \E_{S, \mc{R}} \Big[ \frac{1}{n}\sum^n_{i=1} f(A_{\mc{R}} (S); z_i ) \Big] \Big| \\ 
   & =  \Big|\frac{1}{n}\sum^n_{i=1} \E_{S,z'_i  ,\mc{R}} [ f(A_{\mc{R}} (S); z'_i )] - \E_{S, S', \mc{R}} \Big[ \frac{1}{n}\sum^n_{i=1} f(A_{\mc{R}} (S^{(i)}); z'_i ) \Big] \Big|\label{eq:relabelling}\numberthis \\
   & = \Big| \frac{1}{n}\sum^n_{i=1} \E_{S,S'  ,\mc{R}} [ f(A_{\mc{R}} (S); z'_i )] - \frac{1}{n}\sum^n_{i=1} \E_{S, S', \mc{R}} \left[  f(A_{\mc{R}} (S^{(i)}); z'_i ) \right] \Big|\numberthis\label{eq:gen_hardt_}\\
   & \leq  \frac{1}{n}\sum^n_{i=1} \E_{S,S'  ,\mc{R}} \left[ |  f(A_{\mc{R}} (S); z'_i ) - f(A_{\mc{R}} (S^{(i)}); z'_i )|  \right]\numberthis \label{eq:Hard:missed21} \\
    & \leq 
     L \E_{S,S'  ,\mc{R}} \left[ \frac{1}{n}\sum^n_{i=1} \Vert  A_{\mc{R}} (S) - A_{\mc{R}} (S^{(i)})  \Vert  \right].\numberthis \label{eq:Hard:missed}
\end{align*} 
While \eqref{eq:gen_hardt_} and \eqref{eq:Hard:missed} have been considered in prior works~\cite{hardt2016train,pmlr-v119-lei20c}, we derive uniform upper bounds on the average stability $\frac{1}{n}\sum^n_{i=1} \Vert  A_{\mc{R}} (S) - A_{\mc{R}} (S^{(i)})  \Vert$ for any batch selection rule $\mc{R}$, thus extending generalization error bounds for the SGD to the general class of algorithms $\setalgos$.

\subsection{Minimax Lower Bounds on Algorithmic Generalization}\label{minimax_sec}
Conversely, we establish minimax lower bounds within the class of algorithms $\setalgos$ for all cases of convex, nonconvex and strongly-convex losses. In fact, we see the generalization error as a function of a learning problem $(f,\mc{D})$ in a certain class $\mc{C}$ and an algorithm $A_{\mc{R}}$ in the set $\setalgos$, namely $\gen (f,\mc{D},A_\mc{R} )$. Our purpose then is to discover minimax bounds on the generalization error $\gen (f,\mc{D},A_\mc{R} )$: 
\begin{align}\nonumber
    \sup_{(f,\mc{D})\in \mc{C}} |\gen (f,\mc{D},A_\mc{R} )| \geq  \boxed{\boldsymbol{?}}\, ,\,\,\, \forall A_\mc{R}\in \setalgos 
    \iff
    \inf_{A_\mc{R}\in \setalgos} \sup_{(f,\mc{D})\in \mc{C}} |\gen (f,\mc{D},A_\mc{R} )| \geq \boxed{\boldsymbol{?}}\, ,
\end{align} where the quantity $\boxed{\boldsymbol{?}}$ is independent of the choice of $A_\mc{R}\in \setalgos$. Provided matching upper bounds, in this way we show tightness and of the generalization error within a large class of learning problems and algorithms. Specifically, for Lipschitz smooth convex losses we prove that any algorithm within the class $\setalgos$ is optimal and achieves rates identical to standard rates for the SGD as appeared in prior work~\cite{hardt2016train}. For Lipschitz and smooth nonconvex losses we establish an upper bound which also matches~\cite{hardt2016train} but holds uniformly over all algorithms in $\setalgos$, as well as a lower bound for the more general class of smooth nonconvex losses. In the latter case, our minimax lower bound almost matches upper bounds of full-batch GD from prior work~\cite{nikolakakis2023beyond} showing that in fact full-batch GD is essentially optimal within this class of learning problems and algorithms. Lastly, for strongly-convex losses we prove that all algorithms in $\setalgos$ are optimal within the class of learning problems involving \textit{path-Lipschitz} smooth losses (see Section \ref{sclsection} for precise definitions).

\section{Uniformity with Respect to the Selection Rule}
We start by proving two key properties (Lemma \ref{lemma:table} and Lemma \ref{lemma:Mini-Batch SGD Growth Recursion_}) related to on-average stability, which hold uniformly with respect to any randomized or deterministic batch selection rule $\mc{R}$ (Definition \ref{eq:set_of_algos}). Let $J_t $ be the mini-batch at time $t$ from an input sequence $S\triangleq (z_1 , z_2 ,\ldots ,z_n )$ for some arbitrarily chosen sequence of indices $\{k^t_1,k^t_2,\ldots,k^t_m\}\subseteq [n]$, and $J^{(i)}_t $ be the mini-batch at time $t$ from an input sequence $S^{(i)}\triangleq (z_1 , z_2 ,\ldots, z'_i, \ldots ,z_n )$ for the same arbitrarily chosen sequence of indices $\{k^t_1,k^t_2,\ldots,k^t_m\}$. The first result is on counting the number of different batches $J^{(i)}_t$ compared to $J_t$ for $i\in\{1,\ldots , n\}$. The proof is based on the observation that for any (randomized) data-independent selection rule of mini-batch with size $m$ and each time $t\leq T$, at most $m$ out of $n$ batches $J^{(i)}_t$ ($i\in \{1,\ldots,n\}$) differ with $J_t$ (at one example). 

Specifically, for the rest of the paper we use the notation $\{ J_t \neq J^{(i)}_t\}$ to denote the event where the selection rule chooses the $i^\text{th}$ entry of the respective data-set to be included in the mini-batch (the mini-batches $ J_t$ and $ J^{(i)}_t$ contain the i.i.d. random variables $z_i$ and $z'_i$ at time $t$, respectively\footnote{Although the probability of the event $\{ z_i =z'_i\}$ is not zero in general, we use the notation $\{ J_t \neq J'_t\}$ suggestively. We handle the event $\{ z_i =z'_i\}$ in the analysis when necessary; see also Lemma \ref{lemma:Mini-Batch SGD Growth Recursion_} and the proof of Lemma \ref{lemma:Mini-Batch SGD Growth Recursion_}.}), and $\{ J_t \equiv J^{(i)}_t\}$ to denote the complementary event, i.e., when the selection rule does not choose the $i^\text{ith}$ entry at time $t$. The formal statement of the first lemma follows.
\begin{lemma}\label{lemma:table} Let $\mc{R}$ be any (randomized) selection rule of a sequence of mini-batches with size $m$, namely $\{ \mc{R}_t \}^T_{t=1}$ (independent of the input data-set). Let $J_t $ be the mini-batch at time $t$ from an input sequence $S\triangleq (z_1 , z_2 ,\ldots ,z_n )$ for some arbitrarily chosen indices $\{k^t_1,k^t_2,\ldots,k^t_m\}\subseteq [n]$, and $J^{(i)}_t $ be the mini-batch at time $t$ from an input sequence $S^{(i)}\triangleq (z_1 , z_2 ,\ldots, z'_i, \ldots ,z_n )$ for the same indices with those of $J_t $ namely $\{k^t_1,k^t_2,\ldots,k^t_m\}$. Then 
    \begin{align}
         \sum^n_{i=1} \mathds{1}_{J_t \neq J^{(i)}_t } = m,\quad\text{with probability }1 .
    \end{align}
\end{lemma}
\paragraph{Proof of Lemma \ref{lemma:table}}
The selection rule at time $t$ corresponds to some selection of $m$ distinct indices from the set of integers $\{1,2,\ldots, n \}\triangleq [n] $. Let that choice be $\{k_1 , k_2 ,\ldots, k_m \}\triangleq [K]_m$, then notice that $\mathds{1}_{J_t \neq J^{(i)}_t }=1$ if and only if $i\in [K]_m$, and $\mathds{1}_{J_t \neq J^{(i)}_t }=0$ if and only if $i\notin [K]_m$. This claim can be directly showed through Table \ref{table}, since $S$ and $S^{(i)}$ differ in exactly one entry for all $i\in [n]$. Further the set $[K]_m$ contains exactly $m$ distinct elements, thus we conclude that $\sum^n_{i=1} \mathds{1}_{J_t \neq J^{(i)}_t } = m$, which completes the proof. \qedwhite

We proceed by extending the growth recursion of the SGD with mini-batch size $1$~\cite[Lemma 2.4]{hardt2016train} to the generalized batch selection set of algorithms $\setalgos$ and for any mini-batch of size $m$.
\begin{table}[]
\centering
\begin{tabular}{|l||l|l|l|l|l|l|l|l|l|l|}
\hline
$S$         & $z_1$  & $z_2$  & $z_3$  & $\cdots$ & $z_{i-1}$ & $z_i$  & $z_{i+1}$ & $\cdots$ & $z_{i-1}$    & $z_n$   
\\ \hline \hline
$S^{(1)}$   & $\color{red}z'_1$ & $z_2$  & $z_3$  & $\cdots$ & $z_{i-1}$ & $z_i$  & $z_{i+1}$ & $\cdots$ & $z_{i-1}$    & $z_n$   
\\ \hline
$S^{(2)}$   & $z_1$  & $\color{red}z'_2$ & $z_3$  & $\cdots$ & $z_{i-1}$ & $z_i$  & $z_{i+1}$ & $\cdots$ & $z_{i-1}$    & $z_n$   \\ \hline
$S^{(3)}$   & $z_1$  & $z_2$  & $\color{red}z'_3$ & $\cdots$ &  $z_{i-1}$ & $z_i$  & $z_{i+1}$ & $\cdots$ & $z_{i-1}$    & $z_n$   \\ \hline
$\vdots$     &        &        &        &     $\ddots$    &         &        &         &         &            & $\vdots$ 
\\ \hline
$S^{(i-1)}$   & $z_1$  & $z_2$  & $z_3$  & $\cdots$ & $\color{red}z'_{i-1}$ & $z_i$ & $z_{i+1}$ & $\cdots$ & $z_{i-1}$    & $z_n$   \\ \hline
$S^{(i)}$   & $z_1$  & $z_2$  & $z_3$  & $\cdots$ & $z_{i-1}$ & $\color{red}z'_i$ & $z_{i+1}$ & $\cdots$ & $z_{i-1}$    & $z_n$   \\ \hline
$S^{(i+1)}$   & $z_1$  & $z_2$  & $z_3$  & $\cdots$ & $z_{i-1}$ & $z_{i}$ & $\color{red}z'_{i+1}$ & $\cdots$ & $z_{i-1}$    & $z_n$   \\ \hline
$\vdots$  
&        &        &        &         &         &        &         &   $\ddots$      &            & $\vdots$ 
\\ \hline
$S^{(n-1)}$ & $z_1$  & $z_2$  & $z_3$   & $\cdots$ & $z_{i-1}$ &    $z_i$    &  $z_{i+1}$       & $\cdots$        & $\color{red}z'_{n-1}$ & $z_n$   \\ \hline
$S^{(n)}$   & $z_1$  & $z_2$  & $z_3$  & $\cdots$ & $z_{i-1}$ & $z_i$  & $z_{i+1}$ & $\cdots$ & $z_{n-1}$    & $\color{red}z'_n$  \\ \hline
\end{tabular}
\\[10pt]
\caption{Graphical aid for the proof of Lemma \ref{lemma:table}. Observe that the data-set instances $S$ and $S^{(i)}$ differ in exactly one (at most) entry for all $i\in [n]$.}\label{table}
\end{table}
\subsection{Growth Recursion for Generalized Batch Selection}
\noindent Next, we show the growth recursion lemma similar to~\cite{hardt2016train}, which applies to any algorithm in the set $\setalgos$ (Definition \ref{eq:set_of_algos}) with mini-batch size $m$. We consider a uniform smoothness assumption on the loss with respect to the set of examples, identical to the settings of the prior work~\cite{hardt2016train,pmlr-v119-lei20c,lei2020sharper,pmlr-v178-schliserman22a,taheri2023generalization,pmlr-v206-taheri23a}. 
\begin{assumption}[$\beta$-Smooth Loss]\label{ass:smooth} There exists a constant $\beta\geq 0$, such that for all $w,u\in\mbb{R}^d$ and $z\in \mc{Z}$, $\Vert  \nabla_w f( w, z) - \nabla_u  f( u, z) \Vert_2 \leq \beta \Vert w- u \Vert_2 $.
\end{assumption}  We define the \textit{update map} for any $A_{\mc{R}}\in\setalgos$ at time $t$ as $G_{J_t} (w) \triangleq w-  \frac{\eta_t}{m}\sum_{z\in J_t} \nabla f (w , z)$ (see Algorithm \ref{alg:gen_SGD}). Leveraging the Lipschitzness and smoothness of the loss we may derive a growth recursion that holds universally for all gradient-based algorithms in $\setalgos$ with a mini-batch of size $m$. 
For the case of (necessarily non-Lipschitz) strongly-convex losses we establish the growth recursion seperately in Section \ref{sclsection}, to avoid clutter and keep the exposition concise.
\begin{lemma}[Growth Recursion]\label{lemma:Mini-Batch SGD Growth Recursion_}
Choose an algorithm $A_{\mc{R}}\in \setalgos$, let $w_1 =w^{(i)}_1$ be any common starting point, and set $w_{t+1} = G_{\!J_t} (w_t) $ and $w^{(i)}_{t+1} = G_{\!J^{(i)}_t} (w^{(i)}_t) $, $t\in \{1,\ldots , T\}$. Then for any batch size $m\triangleq |J_t|=|J^{(i)}_t|$, any $t\geq 0$ and any $i\in[n]$ the following recursions hold:
\begin{itemize}
    \item If the loss is convex, then
    \begin{align}
   \!\!\!\!\!\!\!\!\Vert G_{\!J_t} (w_t) - G_{\!J^{(i)}_t} (w^{(i)}_t) \Vert  \leq 
     \begin{cases}
       \Vert w_t - w^{(i)}_t \Vert &\mathrm{under}\text{ } \{J_t \equiv J^{(i)}_t\} \\
         \Vert w_t - w^{(i)}_t \Vert + \frac{2L}{m}  \eta_t  &\mathrm{under}\text{ } \{J_t \neq J^{(i)}_t \}
     \end{cases}.
\end{align}
\item If the loss is nonconvex, then
\begin{align}
   \!\!\!\!\!\!\!\!\Vert G_{\!J_t} (w_t) - G_{\!J^{(i)}_t} (w^{(i)}_t) \Vert  \leq 
     \begin{cases}
       (1 + \beta\eta_t )\Vert w_t - w^{(i)}_t \Vert &\mathrm{under}\text{ } \{J_t \equiv J^{(i)}_t\} \\
       ( 1+ \beta \eta_t )  \Vert w_t - w^{(i)}_t \Vert + \frac{2L}{m}  \eta_t  &\mathrm{under}\text{ } \{J_t \neq J^{(i)}_t\}
     \end{cases}.
\end{align}
\end{itemize} 
\end{lemma}
We prove Lemma \ref{lemma:Mini-Batch SGD Growth Recursion_} in Appendix \ref{growth1proof}. We consider slightly different notation in the statement of the growth recursion compared to~\cite[Lemma 2.5]{hardt2016train} by considering the events $\{J_t \neq J^{(i)}_t\}$ and $\{J_t = J^{(i)}_t\}$ that are complement to each other. This later simplifies the proof of the stability recursion for any $A_{\mc{R}}\in \setalgos$. 
\section{Convex Loss}\label{convexsection}
In this section we derive matching upper and lower generalization error bounds for the class of learning problems with Lipschitz and smooth convex losses, and for all algorithms $A_{\mc{R}}$ in the class $\setalgos$. We start by defining the class of learning problems under consideration.
\begin{definition}[Lipschitz and Smooth Convex Class]\label{defconvexlearnprobm}
    For fixed $L>0$ and $\beta >0$, the class $\mc{C}^{L}_{\beta}$ contains all learning problems $(f,\mc{D})$ with $L$-Lipschitz, $\beta$-smooth, convex losses $f(\cdot,Z)$, $Z\sim \mc{D}$. 
\end{definition}
We proceed by providing the on-average stability guarantees for the learning problems class $\mc{C}^{L}_{\beta}$. Recall that our results hold uniformly over the class of algorithms $\setalgos$.
\begin{theorem}[Generalization Error Upper Bound---Convex Loss]\label{thm:OnAVER_C}
Choose $L>0$ and $\beta>0$. For any learning problem $(f,\mc{D}) \in \mc{C}^{L}_{\beta}$, and for any algorithm $A_{\mc{R}}\in\setalgos$ with step-size $\eta_t< 2/\beta$ and number of iterations $T$, on-average algorithmic stability is bounded as
\begin{align*}
    \frac{1}{n}\sum^n_{i=1} \Vert  A_{\mc{R}} (S) - A_{\mc{R}} (S^{(i)})  \Vert \leq  \frac{2L}{m n}  \sum^{T}_{t=1}  \eta_t \sum^n_{i=1}\mathds{1}_{J_t \neq J^{(i)}_t }
     & = \frac{2L}{n}  \sum^{T}_{t=1}  \eta_t.
\end{align*} Additionally, for any algorithm $A_{\mc{R}}\in\setalgos$ it is true that \begin{align}
     \sup_{\substack{(f,\mc{D})\in \mc{C}^{L}_\beta }} |\gen (f , \mc{D},A_{\mc{R}} )|\leq \frac{2L^2}{n}  \sum^{T}_{t=1}  \eta_t .
\end{align}
\end{theorem}
We proceed with the proof of the theorem.
\paragraph{Proof of Theorem \ref{thm:OnAVER_C}}
Lemma \ref{lemma:Mini-Batch SGD Growth Recursion_} gives
\begin{align*}
    \Vert w_{t+1}- w^{(i)}_{t+1} \Vert &\leq \Vert w_t - w^{(i)}_t \Vert \mathds{1}_{ J_t = J^{(i)}_t  } +   \Vert w_t - w^{(i)}_t \Vert \mathds{1}_{J_t \neq J^{(i)}_t } + \frac{2}{m} L \eta_t \mathds{1}_{J_t \neq J^{(i)}_t }\\
    & = \Vert w_t - w^{(i)}_t \Vert + \frac{2}{m} L \eta_t \mathds{1}_{J_t \neq J^{(i)}_t }.
\end{align*}
By solving the recursion we find
\begin{align}
    \Vert  A_{\mc{R}} (S) - A_{\mc{R}} (S^{(i)})  \Vert \equiv \Vert w_{T+1}- w^{(i)}_{T+1} \Vert &\leq  \frac{2L}{m}  \sum^{T}_{t=1}  \eta_t \mathds{1}_{J_t \neq J^{(i)}_t }  \implies \nonumber\\
     \frac{1}{n}\sum^n_{i=1} \Vert  A_{\mc{R}} (S) - A_{\mc{R}} (S^{(i)})  \Vert &\leq  \frac{2L}{m}  \sum^{T}_{t=1}  \eta_t \lp \frac{1}{n}\sum^n_{i=1}\mathds{1}_{J_t \neq J^{(i)}_t }\rp. \label{eq:...}
\end{align} Inequality \eqref{eq:...} and Lemma \ref{lemma:table} give the on-average stability bound of the statement. This and inequality \eqref{eq:Hard:missed} give the generalization error bound for convex smooth and Lipschitz loss for the final iterate for any batch schedule (see Algorithm \ref{alg:gen_SGD}), completing the proof. \qedwhite

\medskip
Note that Theorem \ref{thm:OnAVER_C} holds for any algorithm $A_\mc{R}\in \setalgos$, and extends established results in prior works for the SGD with batch size $1$~\cite{hardt2016train}. Specifically, the generalization error bound of Theorem \ref{thm:OnAVER_C} holds for general stochastic or deterministic gradient schemes (Definition \ref{eq:set_of_algos}) and any batch size choice $m$. 
Next, we present a matching lower bound on the generalization error for the class of Lipschitz and smooth convex problems $\mc{C}^{L}_{\beta}$, instance-wise relative to the parameters $L$ and $\beta$.

\begin{theorem}[Generalization Error Lower Bound---Convex Loss]\label{Thm:Lower_bound_convex}
Choose constants $L>0$, $\beta >0$, 
step size $\eta_t \leq 1/\beta$, number of iterations $T$, data-set size $n$ and any initial point (independent of the data-set). It is true that
\begin{align*}
  \inf_{A_{\mc{R}}\in \setalgos} \sup_{\substack{(f,\mc{D})\in \mc{C}^{L}_\beta }} |\gen (f , \mc{D},A_{\mc{R}} )| \geq  \frac{L^2}{ 2n} \sum^{T}_{t=1} \eta_t .
\end{align*}
\end{theorem}
To prove Theorem \ref{Thm:Lower_bound_convex}, we consider an $L$-Lipschitz, $\beta$-smooth convex loss parameterized over $L,\beta>0$, and we find an explicit lower bound of the generalization error for all the algorithmic instances $A_{\mc{R}}\in\setalgos$ at the final iteration. We prove Theorem \ref{Thm:Lower_bound_convex} in Appendix \ref{convexlowerproof}. As a consequence of the upper bound of the generalization error in Theorem \ref{thm:OnAVER_C} and the minimax bound in Theorem \ref{Thm:Lower_bound_convex} we conclude that all algorithms in the set $\setalgos$ are \textit{simultaneously optimal} with respect to the learning problem class $\mc{C}^{L}_\beta$.
\section{Nonconvex Loss}\label{nonconvexloss}
Herein, we show generalization error guarantees for learning problems with general Lipschitz and/or smooth (possibly nonconvex) loss functions. Specifically, we consider the class of Lipschitz and smooth losses for the upper bounds of the generalization error. Then we provide minimax generalization error bounds for smooth losses, and we show that in particular \textit{full-batch} GD is essentially optimal within this general class. The definitions of the classes of learning problems considered here is as follows.
\begin{definition}[Lipschitz and Smooth Class]\label{defnconvexlearnprobm}
    For fixed $L >0$ and $\beta >0$, the class $\mc{NC}^{L}_{\beta}$ contains all learning problems $(f,\mc{D})$ with $L$-Lipschitz, $\beta$-smooth non-convex losses $f(\cdot,Z)$, $Z\sim\mc{D}$.
\end{definition}
Of course, for $L=\infty$ we obtain the class $\mc{NC}^{\infty}_{\beta}$ containing all $\beta$-smooth learning problems (with possibly nonconvex loss). We proceed with the corresponding upper bound on the on-average stability and generalization error within the class $\mc{NC}^{L}_{\beta}$. Similarly to the convex case, the bounds apply uniformly over the set of algorithms $\setalgos$.
\begin{theorem}[Generalization Error Upper Bound---Nonconvex Loss]\label{thm:OnAVER_NC}
Choose $L>0$ and $\beta>0$. For any learning problem $(f,\mc{D})\in \mc{NC}^{L}_{\beta}$
and for any algorithm $A_{\mc{R}}\in\setalgos$ with step-size $\eta_t$ and number of iterations $T$, it holds that
\begin{align*}
    \frac{1}{n}\sum^n_{i=1} \Vert  A_{\mc{R}} (S) - A_{\mc{R}} (S^{(i)})  \Vert \leq  
     \frac{2L}{n}  \sum^{T}_{t=1}  \eta_t  \prod^{T}_{j=t+1} (1+\beta \eta_j).
\end{align*}In particular, for any algorithm $A_{\mc{R}}\in\setalgos$ with a decreasing step-size $\eta_t\leq C/t$ where $C<1/\beta$, it is true that 
\begin{align}
    \sup_{\substack{(f,\mc{D})\in \mc{NC}^{L}_{\beta}   } }  |\gen (f , \mc{D},A_{\mc{R}} ) |\leq \frac{2 C e^{C\beta} L^2 T^{C\beta }}{n}  \min\Big\{  1+ \frac{1}{C \beta} , \log(eT) \Big\}.
\end{align}
\end{theorem}
\paragraph{Proof of Theorem \ref{thm:OnAVER_NC}}
Lemma \ref{lemma:Mini-Batch SGD Growth Recursion_} gives \begin{align*}
    \Vert w_{t+1}- w^{(i)}_{t+1} \Vert &\leq (1 + \beta\eta_t )\Vert w_t - w^{(i)}_t \Vert \mathds{1}_{ J_t = J^{(i)}_t  } + \lp 1+ \beta \eta_t \rp  \Vert w_t - w^{(i)}_t \Vert \mathds{1}_{J_t \neq J^{(i)}_t } + \frac{2}{m} L \eta_t \mathds{1}_{J_t \neq J^{(i)}_t }\\
    & \leq (1 + \beta\eta_t )\Vert w_t - w^{(i)}_t \Vert + \frac{2}{m} L \eta_t \mathds{1}_{J_t \neq J^{(i)}_t }.
\end{align*}
By solving the recursion we find
\begin{align}
    \Vert  A_{\mc{R}} (S) - A_{\mc{R}} (S^{(i)})  \Vert \equiv \Vert w_{T+1}- w^{(i)}_{T+1} \Vert &\leq  \frac{2L}{m}  \sum^{T}_{t=1}  \eta_t \mathds{1}_{J_t \neq J^{(i)}_t } \prod^{T}_{j=t+1} (1+\beta \eta_j) \implies \nonumber\\
     \frac{1}{n}\sum^n_{i=1} \Vert  A_{\mc{R}} (S) - A_{\mc{R}} (S^{(i)})  \Vert &\leq  \frac{2L}{m}  \sum^{T}_{t=1}  \eta_t \lp \frac{1}{n}\sum^n_{i=1}\mathds{1}_{J_t \neq J^{(i)}_t }\rp \prod^{T}_{j=t+1} (1+\beta \eta_j). \label{eq:stabthenc}
\end{align}
\noindent The inequality \eqref{eq:Hard:missed} together with \eqref{eq:stabthenc} and Lemma \ref{lemma:table} give the generalization error bound for nonconvex smooth and Lipschitz loss for the final iterate of any batch selection (Algorithm \ref{alg:gen_SGD})\begin{align}
    |\gen|\leq \frac{2L^2}{n}  \sum^{T}_{t=1}  \eta_t  \prod^{T}_{j=t+1} (1+\beta \eta_j).
\end{align} Thus, for any learning problem $(f,\mc{D})\in \mc{NC}^{L}_{\beta}$ and for any algorithm $A_{\mc{R}}\in\setalgos$ with a decreasing step size $\eta_t\leq C/t$ (fixed $C<1/\beta$), it is true that \cite[Lemma 15]{nikolakakis2023beyond} \begin{align}
    \sum^T_{t=1} \eta_t \prod^T_{j=t+1} \lp 1 +  \beta \eta_j \rp \leq   C e^{C\beta} T^{C\beta }\min\left\{  1+ \frac{1}{C \beta} , \log(eT) \right\},
    \end{align} and \begin{align}
    |\gen|\leq \frac{2 C e^{C\beta} L^2 T^{C\beta }}{n}  \min\left\{  1+ \frac{1}{C \beta} , \log(eT) \right\}.
\end{align}
 Inequality \eqref{eq:stabthenc} and Lemma \ref{lemma:table} complete the proof.\qedwhite
 \medskip
 
The next theorem provides a minimax lower bound on the generalization error for learning problems $(f,\mc{D})$ within the class $\mc{NC}^{\infty}_{\beta}$ (Definition \ref{defnconvexlearnprobm}) (smooth and possibly nonconvex losses).
\begin{theorem}[Generalization Error Lower Bound---Nonconvex Loss]\label{Thm:Lower_bound_nonconvex}
Choose $\beta >0$, step size $\eta_t =c/\beta t\leq 1/\beta$, number of iterations $T$, data-set size $n$ and any initial point (independent of the data-set)
. Then, it is true that \begin{align}
   \inf_{A_{\mc{R}}\in \setalgos} \sup_{\substack{(f,\mc{D})\in \mc{C}^{\infty}_\beta }} |\gen (f , \mc{D},A_{\mc{R}} )| 
   & \geq   
   \frac{1    }{2n  }   
   \Big( (T+1)^{\log(1+c)}
   -1 \Big).
\end{align}
\end{theorem}
We refer the reader to Appendix \ref{nonconvex_proof_lower} for the proof of Theorem \ref{Thm:Lower_bound_nonconvex}. The class of learning problems $\mc{NC}^{\infty}_\beta$ in Theorem \ref{Thm:Lower_bound_nonconvex} is larger than the class $\mc{NC}^{L}_\beta$ of the upper bound in Theorem \ref{thm:OnAVER_NC}. {We proceed with a comparison between the lower bound of Theorem \ref{Thm:Lower_bound_nonconvex} and an upper bound on the generalization error from prior work.}
\begin{remark}[{Optimality of full-batch GD}]
We may show that the full-batch GD is almost optimal (up to a $\sqrt{\log T}$ factor and slightly different order of the root) for general smooth losses within the set of algorithms $\setalgos$. 
To explain this further, we compare upper bounds of the generalization error that appear in prior work~\cite[Corollary 8]{nikolakakis2023beyond}, where it has been shown that the generalization error of full-batch GD is \begin{align}
    \sup_{\substack{(f,\mc{D})\in \mc{NC}^{\infty}_\beta}}|\gen (f , \mc{D}, \mathrm{FB\,\, GD} )|= \mc{O} \lp \frac{ T^c\sqrt{\log (T)}}{n} \rp.\label{eq:upper_beyond}
\end{align} By comparing the minimax bound in Theorem \ref{Thm:Lower_bound_nonconvex} with the upper bound in prior work (inequality \eqref{eq:upper_beyond}) we observe the near-optimallity of full-batch GD within the general class of learning problems $\mc{NC}^{\infty}_\beta$. The gap between the roots of the upper and lower bound ($T^{c}$ and $T^{\log(1+c)}$ respectively) can be effectively closed for sufficiently small values of $c$, since it is clear that $\log(1+c)\approx c$ near the origin (for instance, we have the inequality $\log(1+c)\ge c/(c+1)$ is valid for all $c\in[0,1]$) or, maybe more concretely, $\log(1+c)= c + o(c^2)$, as $c\rightarrow 0$. Regardless, our lower bound holds for any choice of $c\in(0,1]$ and provides a generic representation for a wide set of step-size choices. 
\end{remark}
Lastly, we observe that  the generalization error achieved by any algorithm in $\setalgos$ --such as (mini-batch) SGD-- must necessarily be arbitrarily close to that of full-batch GD (also in $\setalgos$) for all sufficiently small stepsizes (possibly up to inconsequential $\sqrt{\log}$ factors). This once again rigorously shows  that stochastic (randomized) training \textit{may} essentially be \textit{at most} as optimal as nonstochastic training, e.g. full-batch training.
\section{Strongly-Convex Loss}\label{sclsection}
For strongly-convex losses we relax the Lipschitz assumption on the loss, since strong-convexity and Lipschitzness are in general incompatible. To do this we consider the maximum norm of gradient loss along the optimization path uniformly over the the data-set, and all algorithms in the given set $\setalgos$. Let $S$ be our data-set, and let $\tilde{S}$ be a another data-set which differs from $S$ in exactly one entry;  let that entry of $\tilde{S}$ be $\tilde{z}$. For simplicity we denote the set of all such data-set pairs as tuples $(S,\tilde{S})$.
Consider the set of algorithms $\setalgos$, as specified in Definition \ref{alg:gen_SGD}. 
We now define the constants
\begin{align*}
      \Tilde{L}^{(1)}_{\eta_t,T}(f)  \triangleq  & \sup_{\substack{(S,\tilde{S}),\,\tilde{c}\in (0,1)\\ A_{\mc{R}}\in \setalgos }} \Vert \nabla f( \tilde{c} A_{\mc{R}}(S) +(1-\tilde{c})A_{\mc{R}}(\tilde{S}),\tilde{z}) \Vert,    \\
       \Tilde{L}^{(2)}_{\eta_t,T}(f) \triangleq         &  \sup_{\substack{t\le T, \,(S,\tilde{S})\\A_{\mc{R}}\in \setalgos}}\Vert \nabla f(w_t , \tilde{z}) \Vert  ,\label{eq:Lf}\numberthis  
    \\
    \Lf \triangleq & \max \Big\{  \Tilde{L}^{(1)}_{\eta_t,T}(f),    \Tilde{L}^{(2)}_{\eta_t,T}(f) \Big\}.
\end{align*} 
We note that the term $\Lf$ essentially places an upper bound on the norm of gradients along any path and any instance of the set of algorithms $\setalgos$. As a consequence, $\Lf$ is determined uniformly over the set $\setalgos$ and the domain $\mc{Z}^{n+1}$. In this way we impose a Lipschitz-like assumption, while avoiding the incompatibility of Lipschitz assumption for general strongly-convex losses.
In what follows, we consider the class of learning problems with strongly-convex and smooth losses, such that the term $\Lf$ is uniformly bounded by some prescribed number. The formal definition follows. \begin{definition}[Smooth \& Strongly-Convex Class]
    Fix an iteration number $T$ and a stepsize sequence $\eta_t$. Choose $\tilde{L}>0, \beta> \gamma>0$. The class $\mc{SC}^{\Tilde{L}}_{\beta,\gamma}$ contains all learning problems $(f,\mc{D})$ with $\beta$-smooth, $\gamma$-strongly-convex losses $f(\cdot,Z),Z\sim \mc{D}$ such that $\Lf \leq \tilde{L}$ .
\end{definition}

In words, the class $\mc{SC}^{\Tilde{L}}_{\beta,\gamma}$ contains all ($d$-dimensional) $\beta$-smooth, $\gamma$-strongly-convex problems which are \textit{uniformly gradient-stable} over the class of algorithms of interest $\setalgos$. We now derive an upper bound on the generalization error by utilizing on-average stability. To begin with, note that for any pair of points $A_{\mc{R}} (S),A_{\mc{R}} (S^{(i)})$, the mean-value theorem  implies the existence of some $\tilde{c}\in [0,1]$ such that  \begin{align*}
    |  f(A_{\mc{R}} (S); z'_i ) - f(A_{\mc{R}} (S^{(i)}); z'_i )|&= |  ( A_{\mc{R}} (S) - A_{\mc{R}} (S^{(i)}) ) \cdot \nabla f( \tilde{c} A_{\mc{R}}(S) +(1-\tilde{c})A_{\mc{R}}(S^{(i)}),z'_i)| \\ &\leq \Vert A_{\mc{R}} (S) - A_{\mc{R}} (S^{(i)}) \Vert  \Vert \nabla f( \tilde{c} A_{\mc{R}}(S) +(1-\tilde{c})A_{\mc{R}}(S^{(i)}),z'_i) \Vert \\
    &\leq \Vert A_{\mc{R}} (S) - A_{\mc{R}} (S^{(i)}) \Vert \Tilde{L}^{(1)}_{\eta_t,T}(f)\\
    &\leq \Tilde{L}_{\eta_t,T} (f) \Vert A_{\mc{R}} (S) - A_{\mc{R}} (S^{(i)}) \Vert.\numberthis \label{eq:MVT}
\end{align*} Similarly to \eqref{eq:Hard:missed21} we obtain the bound \begin{align*}
   |\gen | 
   & \leq  \frac{1}{n}\sum^n_{i=1} \E_{S,S'  ,\mc{R}} \left[ |  f(A_{\mc{R}} (S); z'_i ) - f(A_{\mc{R}} (S^{(i)}); z'_i )|  \right]\\
   &\leq \Lf \E_{S,S'  ,\mc{R}} \left[ \frac{1}{n}\sum^n_{i=1} \Vert  A_{\mc{R}} (S) - A_{\mc{R}} (S^{(i)})  \Vert  \right],\numberthis \label{eq:Hard:missed2}
\end{align*} where \eqref{eq:MVT} gives inequality \eqref{eq:Hard:missed2}. Therefore, for any choice of $(f,\mc{D})\in\mc{SC}^{\Tilde{L}}_{\beta,\gamma}$ \eqref{eq:Hard:missed2} gives \begin{align}
    |\gen | 
   &\leq \tilde{L} \E_{S,S'  ,\mc{R}} \left[ \frac{1}{n}\sum^n_{i=1} \Vert  A_{\mc{R}} (S) - A_{\mc{R}} (S^{(i)})  \Vert  \right].\numberthis \label{eq:LSC}
\end{align} We use \eqref{eq:LSC} as an alternative of \eqref{eq:Hard:missed} to derive bounds on the generalization error in the strongly-convex case. To find upper bounds for the on-average stability term $\frac{1}{n}\sum^n_{i=1} \Vert  A_{\mc{R}} (S) - A_{\mc{R}} (S^{(i)})  \Vert$ in particular, we develop a growth-recursion for strongly-convex losses. 
\begin{lemma}[Growth Recursion---Strongly-Convex Loss]\label{lemma:Mini-Batch sc growth}
Choose $A_{\mc{R}}\in \setalgos$, let $w_1 =w^{(i)}_1$ be the starting point, $w_{t+1} = G_{\!J_t} (w_t) $ and $w^{(i)}_{t+1} = G_{\!J^{(i)}_t} (w^{(i)}_t) $ for any $t\in \{1,\ldots , T\}$. Then for any batch size $m\triangleq |J_t|=|J^{(i)}_t|$, any $t\geq 0$ and any $i\in\{1,\ldots,n\}$, it is true that
\begin{align}
   &\!\!\!\!\!\!\!\! \big\Vert G_{\!J_t} (w_t) - G_{\!J^{(i)}_t} (w^{(i)}_t) \big\Vert \leq \begin{cases}
       (1 - \frac{\eta_t \gamma}{2} )\Vert w_t - w^{(i)}_t \Vert &\mathrm{under}\text{ } \{J_t =J^{(i)}_t\} \\
       (1 - \frac{\eta_t \gamma}{2} )\Vert w_t - w^{(i)}_t \Vert +\frac{2\tilde{L}}{m} \eta_t &\mathrm{under}\text{ } \{ J_t \neq J^{(i)}_t \}
     \end{cases}.
\end{align}
\end{lemma}
We prove Lemma \ref{lemma:Mini-Batch sc growth} in Appendix \ref{eq:proof_growth_strongly}. Through the growth recursion lemma, we derive uniform upper bounds for the on-average stability and the generalization error.
\begin{theorem}[Generalization Error Upper Bound---Strongly-Convex Loss]\label{thm:OnAVER_SC}
  Choose $\beta>0, \gamma>0$ and $\tilde{L}>0$. For any learning problem $(f,\mc{D})\in \mc{SC}^{\Tilde{L}}_{\beta,\gamma}$,  
  and for any algorithm $A_{\mc{R}}\in\setalgos$ with step-size $\eta_t \leq 2/(\beta+\gamma)$ and number of iterations $T$, it holds that
\begin{align*}
    \frac{1}{n}\sum^n_{i=1} \Vert  A_{\mc{R}} (S) - A_{\mc{R}} (S^{(i)})  \Vert \leq  
    \frac{2\tilde{L}}{ n}   \sum^{T}_{t=1}  \eta_t  \prod^{T}_{j=t+1} \bigg(1 - \frac{\eta_j \gamma}{2} \bigg).
\end{align*} 
In particular, if $\eta_t=C \leq 1/(\beta+\gamma)$, then for any algorithm $A_{\mc{R}}\in\setalgos$
\begin{align}
        \sup_{\substack{(f,\mc{D})\in \mc{SC}^{\tilde{L}}_{\beta,\gamma}} } |\gen (f , \mc{D},A_{\mc{R}} ) |\leq \frac{4\tilde{L}^2}{n \gamma} \lp 1-\lp 1-   \frac{C \gamma}{2} \rp^{T} \rp \leq \frac{4 \tilde{L}^2}{n \gamma }.
    \end{align}
\end{theorem}
\paragraph{Proof of Theorem \ref{thm:OnAVER_SC}}
Lemma \ref{lemma:Mini-Batch sc growth} gives \begin{align*}
    \Vert w_{t+1}- w^{(i)}_{t+1} \Vert &\leq \bigg(1 - \frac{\eta_j \gamma}{2} \bigg)\Vert w_t - w^{(i)}_t \Vert \mathds{1}_{ J_t = J^{(i)}_t  } +   \bigg(1 - \frac{\eta_j \gamma}{2} \bigg)\Vert w_t - w^{(i)}_t \Vert \mathds{1}_{J_t \neq J^{(i)}_t } + \frac{2}{m} \tilde{L} \eta_t \mathds{1}_{J_t \neq J^{(i)}_t }\\
    & = \bigg(1 - \frac{\eta_j \gamma}{2} \bigg)\Vert w_t - w^{(i)}_t \Vert + \frac{2}{m} \tilde{L} \eta_t \mathds{1}_{J_t \neq J^{(i)}_t }.
\end{align*}
By solving the recursion we find
\begin{align}
    \Vert  A_{\mc{R}} (S) - A_{\mc{R}} (S^{(i)})  \Vert \equiv \Vert w_{T+1}- w^{(i)}_{T+1} \Vert &\leq  \frac{2\tilde{L}}{m }  \sum^{T}_{t=1}  \eta_t \mathds{1}_{J_t \neq J^{(i)}_t } \prod^{T}_{j=t+1} \bigg(1 - \frac{\eta_j \gamma}{2} \bigg)  \implies \nonumber\\
     \frac{1}{n}\sum^n_{i=1} \Vert  A_{\mc{R}} (S) - A_{\mc{R}} (S^{(i)})  \Vert &\leq  \frac{2\tilde{L}}{m}  \sum^{T}_{t=1}  \eta_t \lp \frac{1}{n}\sum^n_{i=1}\mathds{1}_{J_t \neq J^{(i)}_t }\rp \prod^{T}_{j=t+1} \bigg(1 - \frac{\eta_j \gamma}{2} \bigg) . \label{eq:....}
\end{align} For a fixed step-size choice  $\eta_t = C \leq 1/(\beta+\gamma)$, it is true that~\cite[Lemma 15]{nikolakakis2023beyond}\begin{align}\label{eq:stepsizeequality}
        \sum^{T}_{t=1} \eta_t  \prod^T_{j=t+1}\lp 1-  \frac{\eta_j \gamma}{2} \rp  =  2\frac{1-\lp 1-    \frac{C \gamma}{2} \rp^{T}}{\gamma}.
    \end{align}  The inequalities \eqref{eq:....}, \eqref{eq:stepsizeequality} and Lemma \ref{lemma:table} complete the proof. \qedwhite
\medskip
\\
Lastly, we establish a minimax bound over the set of learning problems of strongly-convex problems in $\mc{SC}^{\Tilde{L}}_{\beta,\gamma}$. The result follows. 
\begin{theorem}[Generalization Error Lower Bound---Strongly-Convex Loss]\label{Thm:Lower_bound}
Choose $\tilde{L}>0$, $\beta\geq \gamma >0$, dimension $d \geq (\beta^2 -\gamma^2) /3 \gamma^2$, number of iterations $T$,
step-size $\eta_t =\eta \in [ 2/\gamma (T+1) , 1/(\beta+\gamma) ]$, input data-set size $n$ and initial point $w_1=0$. Then 
\begin{align}
   \inf_{A_{\mc{R}}\in \setalgos}
   \sup_{\substack{(f, \mc{D})\in \mc{SC}^{\tilde{L}}_{\beta,\gamma} } }  |\gen (f , \mc{D},A_{\mc{R}} ) |\geq \frac{\tilde{L}^2}{32\gamma n}.
\end{align}
\end{theorem}
We compare the bounds of Theorems \ref{thm:OnAVER_SC} and \ref{Thm:Lower_bound}, and we conclude that all algorithms within the set $\setalgos$ are optimal for learning problems in the class $\mc{SC}^{\tilde{L}}_{\beta,\gamma}$. The proof of Theorem \ref{Thm:Lower_bound} appears in Appendix \ref{proof_strongly_lower}.

\section{Conclusion}
We developed upper and lower generalization error bounds for various classes of learning problems, which hold uniformly over mini-batch gradient descent algorithms with data independent batch schedules. In particular, we proved that all gradient based algorithms in the set $\setalgos$ described by Algorithm \ref{alg:gen_SGD} (Definition \ref{eq:set_of_algos}) generalize optimally within the class of Lipschitz smooth convex and strongly-convex losses. For non-convex losses, we showed that full-batch GD is in fact optimal within the class of smooth (possibly nonconvex) losses. Extension of the latter result to the class of Lipschitz and smooth losses, 
 or optimality of other mini-batch schemes, for instance stochastic training schemes such as the SGD algorithm, for smooth (non-Lipschitz) losses remain open problems for future work. An important implication of our results is that different mini-batch schedules mainly affect the optimization error of Lipschitz and smooth losses. Therefore, optimization error guarantees suffice for optimal batch selection in this class of loss functions.

\section{Acknowledgement}
We would like to thank Nathan Srebro for the helpful discussion and valuable suggestions during the development of this work. 
\bibliographystyle{plainurl}
\bibliography{bibliography}
\newpage
\appendix
\section{Convex and Nonconvex Loss}
Herein, we prove the growth recursion (Lemma \ref{lemma:Mini-Batch SGD Growth Recursion_}), and minimax bounds for convex and nonconvex losses as appear in Section \ref{convexsection} and Section \ref{nonconvexloss}.
\subsection{Proof of Lemma \ref{lemma:Mini-Batch SGD Growth Recursion_} (Growth Recursion)}\label{growth1proof}
For two input sequences $S,S^{(i)}$ which differ in the $i^{\text{th}}$ element the update rules are
\begin{align*}
    G_{\!J_t}(w) \triangleq  w -  \frac{\eta_t}{m} \sum_{z \in J_t} \nabla f(w ,z ) ,\,\,\,
    G_{\!J^{(i)}_t}(w) \triangleq  w - \frac{\eta_t}{m} \sum_{z \in J^{(i)}_t} \nabla f(w ,z ),   \numberthis
\end{align*} for potentially different batches $J_t\triangleq J_t (S)$ and $J^{(i)}_t \triangleq J^{(i)}_t (S^{(i)})$ (under the same selection rule), respectively. For nonconvex losses, under the event $\{J_t \equiv J^{(i)}_t \}$
\begin{align*}
    &\Vert G_{\!J_t}(w_t) - G_{\!J_t}(w^{(i)}_t) \Vert \\ 
    & \leq \Vert   w_t -  \frac{\eta_t}{m} \sum_{z \in J_t} \nabla f(w ,z )|_{w=w_t}     -   w^{(i)}_t - \frac{\eta_t}{m} \sum_{z \in J^{(i)}_t} \nabla f(w ,z )|_{w=w^{(i)}_t}   \Vert \\
    &\leq \Vert w_t - w^{(i)}_t \Vert + \frac{\eta_t}{m} \bigg\Vert \sum_{z \in J_t} \nabla_w f(w ,z )|_{w=w_t} -  \sum_{z \in J_t} \nabla_w f(w ,z )|_{w=w^{(i)}_t} \bigg\Vert\\
    &\leq \Vert w_t - w^{(i)}_t \Vert + \frac{\eta_t}{m} \sum_{z \in J_t}\bigg\Vert \nabla_w f(w ,z )|_{w=w_t} -   \nabla_w f(w ,z )|_{w=w^{(i)}_t} \bigg\Vert\\
    &\leq  \Vert w_t - w^{(i)}_t \Vert + \frac{\eta_t}{m} \sum_{z \in J_t} \beta \Vert w_t - w^{(i)}_t \Vert\\
    & = (1 + \beta\eta_t )\Vert w_t - w^{(i)}_t \Vert.\numberthis \label{eq:expand_nonconvex}
\end{align*} 

\noindent Under the event $J_t \neq J^{(i)}_t$, define $J^{-i}_t \triangleq J_t\setminus\{ z_{i} \}$ and $J'^{-i}_t \triangleq J^{(i)}_t\setminus\{ z'_i \}$, and notice that $J^{-i}_t = J'^{-i}_t$ for any $t\leq T$ w.p. $1$. Thus we may decompose the growth recursion as follows \begin{align*}
    &\Vert G_{\!J_t}(w_t) - G_{\!J^{(i)}_t}(w^{(i)}_t) \Vert \\
    & \leq \Vert   w_t -  \frac{\eta_t}{m} \sum_{z \in J_t} \nabla f(w ,z )|_{w=w_t}     -   w^{(i)}_t -  \frac{\eta_t}{m} \sum_{z \in J^{(i)}_t} \nabla f(w ,z )|_{w=w^{(i)}_t}   \Vert 
    \\&= \bigg\Vert w_t - w^{(i)}_t - \frac{\eta_t}{m} \sum_{z \in J_t} \nabla_w f(w ,z )|_{w=w_t} + \frac{\eta_t}{m} \sum_{z' \in J^{(i)}_t} \nabla_w f(w ,z' )|_{w=w^{(i)}_t} \bigg\Vert\\
    &= \bigg\Vert \frac{1}{m} \sum_{z \in J^{-i}_t} \lp w_t - \eta_t \nabla_w f(w ,z )|_{w=w_t}   \rp  - \frac{1}{m} \sum_{z' \in J'^{-i}_t} \lp w^{(i)}_t - \eta_t \nabla_w f(w ,z' )|_{w=w^{(i)}_t}   \rp \\
    & \quad + \frac{1}{m} \lp  w_t - \eta_t \nabla_w f(w ,z_i )|_{w=w_t} \rp  - \frac{1}{m} \lp w^{(i)}_t - \eta_t \nabla_w f(w ,z'_i )|_{w=w^{(i)}_t} \rp \bigg\Vert\\
    &= \bigg\Vert \frac{1}{m} \sum_{z \in J^{-i}_t} ( \underbrace{w_t - \eta_t \nabla_w f(w ,z )|_{w=w_t}}_{G(w_t ,z)}  )  - \frac{1}{m} \sum_{z \in J^{-i}_t} (\underbrace{ w^{(i)}_t - \eta_t \nabla_w f(w ,z )|_{w=w^{(i)}_t} }_{G(w^{(i)}_t ,z)}  ) \\
    & \quad + \frac{1}{m} (  \underbrace{w_t - \eta_t \nabla_w f(w ,z_i )|_{w=w_t} }_{G(w_t,z_i)} )  - \frac{1}{m} ( \underbrace{w^{(i)}_t - \eta_t \nabla_w f(w ,z'_i )|_{w=w^{(i)}_t}}_{G(w^{(i)}_t,z'_i)} ) \bigg\Vert\\
    &= \frac{1}{m}\bigg\Vert  \sum_{z \in J^{-i}_t}\lp G(w_t,z)-G(w^{(i)}_t,z)\rp  +  G(w_t,z_i)  -  G(w^{(i)}_t,z'_i) \bigg\Vert\\
    &\leq  \frac{1}{m}\bigg\Vert  \sum_{z \in J^{-i}_t}\lp G(w_t,z)-G(w^{(i)}_t,z)\rp\bigg\Vert + \frac{1}{m}\Vert G(w_t,z_i)  -  G(w^{(i)}_t,z'_i) \Vert\\
    &\leq  \frac{1}{m}\sum_{z \in J^{-i}_t}\Vert   G(w_t,z)-G(w^{(i)}_t,z) \Vert +\frac{1}{m} \Vert G(w_t,z_i)  -  G(w^{(i)}_t,z'_i) \Vert. \label{eq:applyLemma2.5}\numberthis
\end{align*} for nonconvex loss (in a similar way as appears in \cite[Lemma 2.4]{hardt2016train}) gives that for a random variable $z$ it is true that\begin{align}
    \Vert   G(w_t,z)-G(w^{(i)}_t,z) \Vert &\leq (1+\beta \eta_t) \Vert w_t -w^{(i)}_t \Vert  \label{eq:first_corected}
    \end{align} additionally for two i.i.d. random variables $z_i$, $z'_i$ (the probability of $\{z_i =z'_i\}$ is positive for discrete random variables) and
    \begin{align*}
    \Vert  G(w_t,z_i)-G(w^{(i)}_t,z'_i )\Vert &\leq \begin{cases}
        \Vert w_t -w^{(i)}_t \Vert +  2L\eta_t  &\quad \text{if } z_i \neq z'_i ,\\
       (1+\beta \eta_t) \Vert w_t -w^{(i)}_t \Vert &\quad \text{if } z_i = z'_i .
     \end{cases} 
\end{align*} Therefore, for both cases it is true that\begin{align}
    \Vert  G(w_t,z_i)-G(w^{(i)}_t,z'_i )\Vert &\leq 
        (1+\beta \eta_t)\Vert w_t -w^{(i)}_t \Vert +  2L\eta_t  \label{eq:second_corected}
\end{align} By combining the inequalities \eqref{eq:first_corected}, \eqref{eq:second_corected} together with \eqref{eq:applyLemma2.5} we find \begin{align*}
    \Vert G_{\!J_t}(w_t) - G_{\!J^{(i)}_t}(w^{(i)}_t) \Vert &\leq \frac{1}{m}\sum_{z \in J^{-i}_t}(1+\beta \eta_t) \Vert w_t -w^{(i)}_t \Vert +\frac{1}{m} \lp (1+\beta \eta_t) \Vert w_t -w^{(i)}_t \Vert + 2 L \eta_t\rp\\
    &= \frac{m-1}{m}(1+\beta \eta_t) \Vert w_t -w^{(i)}_t \Vert +\frac{1}{m} \lp  (1+\beta \eta_t)\Vert w_t -w^{(i)}_t \Vert + 2 L \eta_t\rp\\
    &= \lp 1+ \beta \eta_t \rp  \Vert w_t -w^{(i)}_t \Vert + \frac{2L}{m}  \eta_t .\numberthis
\end{align*} The last display gives the second part of the recursion and completes the proof for nonconvex losses.
For convex loss \cite[Lemma 2.4]{hardt2016train} gives \begin{align}
    \Vert   G(w_t,z)-G(w^{(i)}_t,z) \Vert &\leq  \Vert w_t -w^{(i)}_t \Vert,\\
    \Vert  G(w_t,z_i)-G(w^{(i)}_t,z'_i )\Vert &\leq  \Vert w_t -w^{(i)}_t \Vert +  \eta_t \lp  \Vert  \nabla f(w_t ,z_i ) \Vert    + \Vert \nabla f(w^{(i)}_t ,z'_i ) \Vert \rp,\label{eq:second_no_corr_needed}
\end{align} and the inequality \eqref{eq:second_no_corr_needed} holds under both events $\{z_i = z'_i \}$ and $\{z_i \neq z'_i \}$. By combining the last two together with \eqref{eq:applyLemma2.5} we find \begin{align*}
    \Vert G_{\!J_t}(w_t) - G_{\!J^{(i)}_t}(w^{(i)}_t) \Vert &\leq \frac{1}{m}\sum_{z \in J^{-i}_t} \Vert w_t -w^{(i)}_t \Vert +\frac{1}{m} \lp  \Vert w_t -w^{(i)}_t \Vert + 2 L \eta_t\rp\\
    &= \frac{m-1}{m} \Vert w_t -w^{(i)}_t \Vert +\frac{1}{m} \lp  \Vert w_t -w^{(i)}_t \Vert + 2 L \eta_t\rp\\
    &=   \Vert w_t -w^{(i)}_t \Vert + \frac{2L}{m}  \eta_t .\numberthis \label{eq:secondpartconvexgr}
\end{align*} Finally, we find the non-expansive recursion for convex loss when $J_t \equiv J^{(i)}_t$,
\begin{align*}
    \Vert G_{\!J_t}(w_t) - G_{\!J_t}(w^{(i)}_t) \Vert^2  &= \Vert w_t - w^{(i)}_t \Vert^2 -2   \frac{\eta_t}{m} \inp{ \sum_{z \in J_t} \nabla f(w_t ,z ) - \sum_{z \in J_t} \nabla f(w^{(i)}_t ,z ) }{w_t -w^{(i)}_t} \\&\qquad+ \eta^2_t \Vert \frac{1}{m}\sum_{z \in J_t} \nabla f(w_t ,z ) - \frac{1}{m}\sum_{z \in J_t} \nabla f(w^{(i)}_t ,z ) \Vert^2 \\
    &\leq \Vert w_t - w^{(i)}_t \Vert^2 -2   \frac{\eta_t}{\beta m^2}\Vert\sum_{z \in J_t}\nabla f(w_t ,z ) -  \sum_{z \in J_t} \nabla f(w^{(i)}_t ,z ) \Vert^2  \\&\qquad + \eta^2_t  \Vert \frac{1}{m} \sum_{z \in J_t}\nabla f(w_t ,z ) - \frac{1}{m} \sum_{z \in J_t} \nabla f(w^{(i)}_t ,z ) \Vert^2\\
    &\leq \Vert w_t - w^{(i)}_t \Vert^2 -\frac{\eta_t}{m^2} (  \frac{2}{\beta }-\eta_t)\Vert\sum_{z \in J_t}\nabla f(w_t ,z ) -  \sum_{z \in J_t} \nabla f(w^{(i)}_t ,z ) \Vert^2 \\
    &\leq  \Vert w_t - w^{(i)}_t \Vert^2.    \numberthis \label{eq:convex,nonexpansive}
\end{align*} For the last chain of inequalities we used the co-coersivity of the function $h(w)\triangleq \sum_{z \in J_t}\nabla f(w ,z )$, for which is true that $\inp{h(w)-h(u)}{w-u}\geq \frac{1}{\beta m}\Vert  h(w)-h(u)\Vert^2$ and $\eta_t<2/\beta$. Inequalities \eqref{eq:secondpartconvexgr} and \eqref{eq:convex,nonexpansive} gives the growth recursion for convex losses and completes the proof.\qedwhite
\subsection{Proof Of Theorem \ref{Thm:Lower_bound_convex} (Lower Bound---Convex Loss)}\label{convexlowerproof}
Let the initial point be any $w_1 \in \mbb{R^d}$ (independent of the data-set) and consider the loss 
\begin{align*}
    f(w,\mathbf{z}) = \sum^{d-1}_{k=1} w^k z^k + \frac{\beta}{2}(w^d - w^d_1 -z^d)^2 \mathds{1}_{|w^d - w^d_1 -z^d|\leq \tau} + \beta\tau (|w^d - w^d_1 -z^d|-\frac{1}{2}\tau)\mathds{1}_{|w^d - w^d_1 -z^d|> \tau},\label{eq:convex_f}
\end{align*}
for some $\tau \leq L/\sqrt{d}\beta $, $\mathbf{z}_i \triangleq (z^1_i, z^2_i ,\ldots ,z^{d}_i )$, such that  $ (z^1_i, z^2_i ,\ldots ,z^{d-1}_i )\in\{ -L/\sqrt{d} ,+L/\sqrt{d} \}^{d-1} $, $z^d_i \in \{ -L/2\beta\sqrt{d} ,+L/2\beta\sqrt{d} \} $  $i\in \{ 1, 2,\ldots , n \}$, $\P (z_i^j = L/\sqrt{d})= 1/2 $ for $j\in\{1,\ldots ,d-1\}$, $\P (z_i^d = L/2\beta \sqrt{d})= 1/2 $ and $z^\ell_k \perp z^{\ell '}_{k'} $ for all distinct pairs $(\ell,k)$, $(\ell' , k')$. Then $f(w,\mathbf{z})$ is $\beta$ smooth and $L$ Lipschitz, and continuously differentiable. For instance, we have
\begin{align}
    &\Vert\nabla f(w,\mathbf{z})\Vert \nonumber\\&= \sqrt{\sum^{d-1}_{k=1} (z^k)^2 + \beta^2 (w^d - w^d_1 -z^d)^2 \mathds{1}_{|w^d - w^d_1 -z^d|\leq \tau} + \beta^2\tau^2 \text{sign}^2(w^d - w^d_1 -z^d) \mathds{1}_{|w^d - w^d_1 -z^d|> \tau}} 
    \nonumber\\ &\leq \sqrt{\sum^{d-1}_{k=1} (z^k)^2 + \beta^2 \tau^2   }\\
    &\leq \sqrt{ \frac{L^2 (d-1)}{d} + \beta^2 \frac{L^2}{d\beta^2 }   }\leq L , \label{eq:lipschitz_convex_lower}
\end{align} 
and smoothness can be verified accordingly. Then the updates with step size $\eta_t  $ and $k\neq d$ are\begin{align}
    w^{k}_{t+1} &=w^k_t -\frac{\eta_t}{m}\sum_{z\in J_t} z^k \implies  w^{k}_{T} = w^k_1 - \sum^{T-1}_{t=1}\frac{\eta_t}{m}\sum_{z\in J_t} z^k,\label{eq:linear_update}
\end{align} 
and for $k=d$ the update follows
\begin{align}
   \!\!\! w^{d}_{t+1} &=w^d_t - \frac{\eta_t}{m}\sum_{z\in J_t}\lp  \beta(w^d_t -w^d_1 -z^d)\mathds{1}_{|w^d_t -w^d_1 -z^d|\leq \tau} + \beta\tau \text{sign} ( w^d_t -w^d_1 -z^d)\mathds{1}_{|w^d_t -w^d_1 -z^d|> \tau} \rp.\label{eq:quadratic_update}
\end{align} Hereafter, choose $\tau=L/(\sqrt{d}\beta)$. It is true that $\eta_t\leq 1/\beta$, and $|z^d|= \tau/2$. Starting from $w_1 $,  and \begin{align}
    w^{d}_{2} &= w^{d}_{1} - \frac{\eta_t}{m}\sum_{z\in J_t}\lp  -\beta z^d\mathds{1}_{| -z^d|\leq \tau} + \beta\tau \text{sign} ( -z^d)\mathds{1}_{|-z^d|> \tau} \rp
    \\&= - \frac{\eta_t}{m}\sum_{z\in J_t}\lp  -\beta z^d\mathds{1}_{| -z^d|\leq \tau}  \rp\implies\nonumber \\
    |w^{d}_{2}-w^{d}_{1} - z^d|& \leq |w^{d}_{2}-w^{d}_{1}| + | z^d|\leq  \tau/2 + \tau/2\leq  \tau .
\end{align} Assuming that $ |w^{d}_{t} -w^{d}_{1}|\leq  \tau/2 $, we can show that  $ |w^{d}_{t+1} -w^{d}_{1}|\leq  \tau/2 $ and $ |w^{d}_{t+1} -w^{d}_{1}- z^d|\leq  \tau $ because \begin{align*}
    &|w^{d}_{t+1} -w^{d}_{1}- z^d| \\&\leq |w^{d}_{t+1}-w^{d}_{1}| + |z^d| \\
    & \leq \bigg|w^d_{t}-w^{d}_{1} - \frac{\eta_t}{m}\sum_{z\in J_t}  \beta (w^d_t -w^d_1 - z^d )\mathds{1}_{|w^d_t -w^d_1 -z^d|\leq \tau} + \beta\tau \text{sign} ( w^d_t -w^d_1-z^d)\mathds{1}_{|w^d_t -w^d_1-z^d|> \tau}  \bigg| +\tau/2 \\
    &= \bigg|(1-\beta \eta_t )(w^d_{t} -w^{d}_{1})- \frac{\eta_t}{m}\sum_{z\in J_t}  \beta ( - z^d )\mathds{1}_{|w^d_{t} -w^{d}_{1}-z^d|\leq \tau} + \beta\tau \text{sign} ( w^d_{t}-w^{d}_{1}-z^d)\mathds{1}_{|w^d_{t}-w^{d}_{1}-z^d|> \tau}  \bigg| +\tau/2\\
    &\leq |(1-\beta \eta_t )(w^d_{t}-w^{d}_{1})| +\bigg| \frac{\eta_t}{m}\sum_{z\in J_t}  \beta  z^d \mathds{1}_{|w^d_{t}-w^{d}_{1} -z^d|\leq \tau} + \beta\tau \text{sign} ( w^d_{t}-w^{d}_{1}-z^d)\mathds{1}_{|w^d_{t}-w^{d}_{1}-z^d|> \tau}  \bigg| +\tau/2 \\
    &=|(1-\beta \eta_t )(w^d_{t}-w^{d}_{1})\Big| +\bigg| \frac{\eta_t}{m}\sum_{z\in J_t}  \beta  z^d   \bigg| +\tau/2 \\
    &\leq \tau/2 +\tau/2 = \tau.
\end{align*} Through the induction above we showed that the updates $w^d_t$ remain in the interval $ |w^{d}_{t} - w^{d}_{1} - z^d|\leq  \tau $ for any $t\leq T$, thus the update in \eqref{eq:quadratic_update} gives \begin{align}
    \!\!\!\!w^{d}_{t+1} &=w^d_t - \frac{ \beta\eta_t}{m}\sum_{z\in J_t} (w^d_t -w^d_1-z^d)
     = (1-\beta \eta_t ) w^d_t + \frac{ \beta\eta_t}{m}\sum_{z\in J_t} (z^d + w^d_1)\\&\implies w^{d}_{T} = w^d_1  \underbrace{\lp \prod^T_{t=1}(1-\beta \eta_t ) + \frac{ \beta}{m}\sum^{T}_{t=1} \eta_t \prod^{T}_{j=t+1} (1-\beta \eta_j )\rp}_{\mc{H}} + \frac{ \beta}{m}\sum^{T}_{t=1} \eta_t\sum_{z\in J_t} z^d \prod^{T}_{j=t+1} (1-\beta \eta_j ).\nonumber
\end{align} Then the output of any algorithm in the set $\setalgos$ for any initial point (independent of the data-set) is \begin{align}
    A_{\mc{R}}^k  (S) &= w^k_1- \sum^{T}_{t=1}\frac{\eta_t}{m}\sum_{z\in J_t} z^k. \quad k\in\{ 1,2,\dots,d-1\} \\
    A_{\mc{R}}^d  (S) &=  w^d_1 \mc{H}+ \frac{ \beta}{m}\sum^{T}_{t=1} \eta_t\sum_{z\in J_t} z^d \prod^{T}_{j=t+1} (1-\beta \eta_j )
\end{align}
Additionally,  \begin{align*}
    \gen & =\E_{S,z,\mc{R}} [f(A_{\mc{R}}  (S); \mathbf{z}) -\frac{1}{n}\sum^n_{i=1} f(A_{\mc{R}}  (S); \mathbf{z}_i )]  \\
    &=   \E_{S,S',\mc{R}} \left[ \frac{1}{n}\sum^n_{i=1} f(A_{\mc{R}}  (S); \mathbf{z}'_i)] -\frac{1}{n}\sum^n_{i=1} f(A_{\mc{R}}  (S); \mathbf{z}_i ) \right] 
\end{align*} We use \eqref{eq:convex_f} and the fact $|A^d_\mc{R} (S)-w^d_1-z^d|\leq \tau$, then \begin{align*}
    &\gen \\& = \E_{S,S',\mc{R}} \left[\frac{1}{n}\sum^n_{i=1} \lp \frac{\beta}{2}( A^d_{\mc{R}}  (S)-w^d_1 -(z_i')^d)^2 -\frac{\beta}{2}( A^d_{\mc{R}}  (S) -w^d_1 -z_i^d)^2 +\sum^{d-1}_{k=1} A^k_{\mc{R}}  (S) ((z_i')^k - z_i^k)   \rp \right]\\
    &= \E_{S,S',\mc{R}} \Bigg[ \frac{1}{n}\sum^n_{i=1} \Bigg(\frac{\beta}{2}(-2 A^d_{\mc{R}}(S)(z_i')^d + 2 A^d_{\mc{R}}(S)z_i^d +((z_i')^d)^2 - (z_i^d )^2 )\\&\qquad\qquad\qquad+2w^d_1 ((z_i')^d) -z_i^d))  +\sum^{d-1}_{k=1} A^k_{\mc{R}}  (S) ((z_i')^k - z_i^k)   \Bigg) \Bigg]\\
    &= \E_{\mc{R}} \left[\frac{1}{n}\sum^n_{i=1} \E_{S}\left[ \frac{\beta}{2}( 2 A^d_{\mc{R}}(S)z_i^d )  +\sum^{d-1}_{k=1} A^k_{\mc{R}}  (S) ( - z_i^k)   \right] \right]\\
    &=\E_{\mc{R}} \left[\frac{1}{n}\sum^n_{i=1} \E_{S}\left[ \beta(  z_i^d \frac{ \beta}{m}\sum^{T}_{t=1} \eta_t\sum_{z\in J_t} z^d \prod^{T}_{j=t+1} (1-\beta \eta_j ) )  +\sum^{d-1}_{k=1}  z_i^k \sum^{T}_{t=1}\frac{\eta_t}{m}\sum_{z\in J_t} z^k  \right] \right],
\end{align*} and for the last two equities we used the fact that the initial point $w_1$ is independent of data-set and that the random variables $z_i ,z'_i$ have zero mean for all $i\leq n$. If $\mathbf{z}_i \in J_t$ then $\E_{S}[ z^d_i \sum_{\mathbf{z}\in J_t } z^d ] =L^2 /\beta^2 d$, otherwise $\E_{S}[ z^d_i \sum_{\mathbf{z}\in J_t } z^k ] =0 $, additionally for $k\in\{1,2,\ldots,d-1\}$, if $\mathbf{z}_i \in J_t$ then $\E_{S}[ z^k_i \sum_{\mathbf{z}\in J_t } z^k ] =L^2 / d$, otherwise $\E_{S}[ z^k_i \sum_{\mathbf{z}\in J_t } z^k ] =0 $  thus $\E_{S}[ z^k_i \sum_{\mathbf{z}\in J_t } z^k ]= \mathds{1}_{\mathbf{z}_i \in J_t} L^2/ d  $ for any (randomized) selection rule, further $\sum^n_{i=1} \mathds{1}_{\mathbf{z}_i \in J_t} = m$ for any $A_{\mc{R}}\in \setalgos$ (Definition \ref{alg:gen_SGD}) with batch-size $m\equiv |J_t|$ for all $t$, and the generalization error is \begin{align*}
    |\gen | & =\left| \E_{\mc{R}} \left[\frac{1}{n}\sum^n_{i=1}  \lp  \frac{ \beta^2}{m}\sum^{T}_{t=1} \eta_t  \prod^{T}_{j=t+1} (1-\beta \eta_j )  \E_{S}[ z_i^d \sum_{z\in J_t} z^d ]  +\sum^{d-1}_{k=1}  \sum^{T}_{t=1}\frac{\eta_t}{m} \E_{S}[ z_i^k \sum_{z\in J_t} z^k ]\rp \right]\right| \\
    & =\left| \E_{\mc{R}} \left[\frac{1}{n}\sum^n_{i=1}  \lp  \frac{ L^2}{dm}\sum^{T}_{t=1} \eta_t  \prod^{T}_{j=t+1} (1-\beta \eta_j )  \mathds{1}_{\mathbf{z}_i \in J_t} +\frac{L^2}{dm}\sum^{d-1}_{k=1}  \sum^{T}_{t=1} \eta_t \mathds{1}_{\mathbf{z}_i \in J_t}\rp \right]\right|\\
    &= \left| \E_{\mc{R}} \left[\frac{1}{n}  \lp  \frac{ L^2}{dm}\sum^{T}_{t=1} \eta_t  \prod^{T}_{j=t+1} (1-\beta \eta_j )  \sum^n_{i=1}\mathds{1}_{\mathbf{z}_i \in J_t} +\frac{L^2}{dm}\sum^{d-1}_{k=1}  \sum^{T}_{t=1} \eta_t \sum^n_{i=1}\mathds{1}_{\mathbf{z}_i \in J_t}\rp \right]\right|\\
    &=  \left| \E_{\mc{R}} \left[\frac{1}{n}  \lp  \frac{ L^2}{d}\sum^{T}_{t=1} \eta_t  \prod^{T}_{j=t+1} (1-\beta \eta_j )   +\frac{L^2}{d}\sum^{d-1}_{k=1}  \sum^{T}_{t=1} \eta_t \rp \right]\right|\\
    &=  \frac{L^2}{d n}  \lp \sum^{T}_{t=1} \eta_t  \prod^{T}_{j=t+1} (1-\beta \eta_j )   +(d-1)  \sum^{T}_{t=1} \eta_t \rp \\
    &\geq  \frac{(d-1)L^2}{d n} \sum^{T}_{t=1} \eta_t \\
    & \geq \frac{L^2}{ 2n} \sum^{T}_{t=1} \eta_t .\label{eq:proof_convex_final}\numberthis
\end{align*} The inequality \eqref{eq:proof_convex_final} holds for any algorithm in the set $\setalgos$, for any choice of step step size $\eta_t$, number of iterations $T$, Lipschitz constant $L$ (due to \eqref{eq:lipschitz_convex_lower})  and completes the proof. \qedwhite
\subsection{Proof of Theorem \ref{Thm:Lower_bound_nonconvex} (Lower Bound---Nonconvex Loss)}\label{nonconvex_proof_lower}
Consider the loss $f(w,\mathbf{z})=  ( w-\mathbf{z})^\T  \Lambda  ( w-\mathbf{z}) /2$, $w\in\mbb{R}^d$, $\Lambda \triangleq \text{diag}(\lambda_1,\ldots,\lambda_d)$, $\mathbf{z}_i =(z^1_i, z^2_i ,\ldots ,z^d_i )\in\{ -1/\sqrt{\beta d} ,+1/\sqrt{ \beta d} \}^{d} $, $i\in \{ 1, 2,\ldots , n \}$, $\P (z_i^j = 1/\sqrt{\beta d})= 1/2 $ and $z^\ell_k \perp z^{\ell '}_{k'} $ for all distinct pairs $(\ell,k)$, $(\ell' , k')$. Then the SGD updates with step size $\eta_t = c/t\beta $ yield \begin{align}
    w_{t+1} &=w_t - \frac{c}{\beta t} \Lambda\lp w_t-\frac{1}{m}\sum_{\mathbf{z}\in J_t } \mathbf{z} \rp = \lp 1- \frac{c}{\beta t} \Lambda \rp w_t +  \frac{c}{m \beta t}\Lambda\sum_{\mathbf{z}\in J_t } \mathbf{z} 
\end{align} 
and for any arbitrary initial point $W_1$ (independent of the data-set)
\begin{align}
    A_{\mc{R}}(S)=w_{T+1} =  \prod^{T}_{t=1}\lp 1- \frac{c}{\beta t} \Lambda \rp w_1 +  \frac{c}{m \beta }\sum^{T}_{t=1}\frac{1}{ t} \prod^{T}_{j=t+1} \lp 1- \frac{c}{\beta j} \Lambda \rp\Lambda\sum_{\mathbf{z}\in J_t } \mathbf{z}.  
\end{align} 
Additionally,  \begin{align*}
    \gen & =\E_{S,z,\mc{R}} [f(A_{\mc{R}}  (S); \mathbf{z}) -\frac{1}{n}\sum^n_{i=1} f(A_{\mc{R}}  (S); \mathbf{z}_i )]  \\
    &=   \E_{S,S',\mc{R}} \left[ \frac{1}{n}\sum^n_{i=1} f(A_{\mc{R}}  (S); \mathbf{z}'_i)] -\frac{1}{n}\sum^n_{i=1} f(A_{\mc{R}}  (S); \mathbf{z}_i ) \right] \\
    &=  \frac{1}{2} \E_{S,S',\mc{R}} \left[ \frac{1}{n}\sum^n_{i=1}  (  A_{\mc{R}} (S)-\mathbf{z}'_i)^\T  \Lambda  (  A_{\mc{R}} (S)-\mathbf{z}'_i) -\frac{1}{n}\sum^n_{i=1} (  A_{\mc{R}} (S)-\mathbf{z}_i)^\T  \Lambda  (  A_{\mc{R}} (S)-\mathbf{z}_i) \right] \\
    & = \frac{1}{2} \E_{S,S',\mc{R}} \Bigg[  \frac{1}{n}\sum^n_{i=1} \big( A^\T_{\mc{R}} (S) \Lambda A_{\mc{R}} (S) - 2A^\T_{\mc{R}} (S) \Lambda \mathbf{z}'_i  + (\mathbf{z}'_i)^\T \Lambda \mathbf{z}'_i \\&\qquad\qquad\qquad\qquad\qquad - A^\T_{\mc{R}} (S) \Lambda A_{\mc{R}} (S)  +  2 A^\T_{\mc{R}} (S) \Lambda  \mathbf{z}_i  - (\mathbf{z}_i)^\T \Lambda \mathbf{z}_i \big) \Bigg]\\
    & =     \frac{1}{2n}\sum^n_{i=1}  - 2\E_{S,\mc{R}}[A^\T_{\mc{R}} (S) ] \Lambda \underbrace{\E_{S'}[\mathbf{z}'_i]}_{=0}  + \underbrace{\E_{S'}[(\mathbf{z}'_i)^\T \Lambda \mathbf{z}'_i ]  - \E_{S}[(\mathbf{z}_i)^\T \Lambda \mathbf{z}_i]}_{=0} +  \E_{S,\mc{R}} [2 A^\T_{\mc{R}} (S) \Lambda \mathbf{z}_i ]   \\
    & =  \frac{1}{2n}\sum^n_{i=1}    \E_{S,\mc{R}} [2 A^\T_{\mc{R}} (S) \Lambda \mathbf{z}_i ]   \\
    & =  \frac{1}{n}\sum^n_{i=1}    \E_{S} \Bigg[ \mathbf{z}^\T_i \Lambda  \frac{c}{m \beta }\sum^{T}_{t=1}\frac{1}{ t} \prod^{T}_{j=t+1} \lp 1- \frac{c}{\beta j} \Lambda \rp\Lambda\sum_{\mathbf{z}\in J_t } \mathbf{z} \Bigg]+\underbrace{\E_{S} [ \mathbf{z}^\T_i]}_{0} \Lambda  \prod^{T}_{t=1}\lp 1- \frac{c}{\beta t} \Lambda \rp \E [ W_1 ]  \\
    & = \frac{c}{m \beta } \sum^{T}_{t=1}\frac{1}{n}\sum^n_{i=1}    \E_{S,\mc{R}} \Bigg[ \mathbf{z}^\T_i \Lambda  \frac{1}{ t} \prod^{T}_{j=t+1} \lp 1- \frac{c}{\beta j} \Lambda \rp\Lambda\sum_{\mathbf{z}\in J_t } \mathbf{z} \Bigg]. \label{eq:nc_1_lb} \numberthis
\end{align*}
Define
\begin{align}
    M^t \triangleq \Lambda  \frac{1}{ t} \prod^{T}_{j=t+1} \lp 1- \frac{c}{\beta j} \Lambda \rp\Lambda.
\end{align}
If $\mathbf{z}_i \in J_t$ then $\E_{S}[ z^k_i \sum_{\mathbf{z}\in J_t } z^k ] =1 /\beta d$, otherwise $\E_{S}[ z^k_i \sum_{\mathbf{z}\in J_t } z^k ] =0 $, thus $\E_{S}[ z^k_i \sum_{\mathbf{z}\in J_t } z^k ]= \mathds{1}_{\mathbf{z}_i \in J_t} 1/\beta d  $ for any (randomized) selection rule, and $\sum^n_{i=1} \mathds{1}_{\mathbf{z}_i \in J_t} = m$ for any $A_{\mc{R}}\in \setalgos$ (Definition \ref{alg:gen_SGD}) with batch-size $m\equiv |J_t|$ for all $t$, thus 
\begin{align*}
    \frac{c}{\beta nm}\E_\mc{R} \left[ \sum^n_{i=1}  \sum^d_{k=1}\sum^T_{t=1} M^t_{kk} \E_{S} \bigg[  z^k_i  \sum_{\mathbf{z}\in J_t } z^k \bigg]     \right] 
    & = \frac{c}{d\beta^2 nm}\E_\mc{R} \left[ \sum^n_{i=1}  \sum^d_{k=1} \sum^T_{t=1} M^t_{kk} \mathds{1}_{\mathbf{z}_i \in J_t} \right] \\
     & = \frac{c}{d \beta^2 nm}\E_\mc{R} \left[  \sum^d_{k=1} \sum^T_{t=1} M^t_{kk} \sum^n_{i=1} \mathds{1}_{\mathbf{z}_i \in J_t}   \right] \\
     & =\frac{c}{d \beta^2 n}\E_\mc{R} \left[  \sum^d_{k=1} \sum^T_{t=1} M^t_{kk}    \right]\\
     & = \frac{c}{d \beta^2 n} \sum^d_{k=1} \sum^T_{t=1} M^t_{kk}   .   \label{eq:nc_2_lb} \numberthis
\end{align*} 
Let $\lambda_k <0$ with  $|\lambda_k|\leq \beta$, and choose $c \le 1$. Then \begin{align*}
    \sum^T_{t=1} M^t_{kk} &= \lambda^2_{k}  \sum^T_{t=1} \frac{1}{ t} \prod^{T}_{j=t+1} \lp 1+ \frac{c}{\beta j}|\lambda_{k}| \rp \\
    & \ge \lambda^2_{k}  \sum^T_{t=1} \frac{1}{ t} \prod^{T}_{j=t+1} 2^{\frac{c|\lambda_k|}{\beta j}} \label{eq:exp_trick} \numberthis \\
    & = \lambda^2_{k}  \sum^T_{t=1} \frac{1}{ t}  2^{\sum^{T}_{j=t+1}\frac{c|\lambda_k|}{\beta j}} \\
    & = \lambda^2_{k}  \sum^T_{t=1} \frac{1}{ t}  2^{\sum^{T}_{j=1}\frac{c|\lambda_k|}{\beta j} -\sum^{t}_{j=1}\frac{c|\lambda_k|}{\beta j}} \\
    & \ge \lambda^2_{k}  \sum^T_{t=1} \frac{1}{ t}  2^{ \frac{c|\lambda_k|}{\beta }\ln (T+1) -\frac{c|\lambda_k|}{\beta }\ln (t)-\frac{c|\lambda_k|}{\beta }} \\
    & =\lambda^2_{k}2^\frac{-c|\lambda_k|}{\beta }  \sum^T_{t=1} \frac{1}{ t}  2^{ \frac{c|\lambda_k|}{\beta }\ln (\frac{T+1}{t}) } \\
    & =\lambda^2_{k}2^\frac{-c|\lambda_k|}{\beta  }  \sum^T_{t=1} \frac{1}{ t}  2^{ \frac{c|\lambda_k|}{\beta \log_2 e}\log_2 (\frac{T+1}{t})  } \\
    & =\lambda^2_{k}2^\frac{-c|\lambda_k|}{\beta  } (T+1)^\frac{c|\lambda_k|}{\beta \log_2 e} \sum^T_{t=1} \frac{1}{ t}   \frac{1}{t^\frac{c|\lambda_k|}{\beta \log_2 e}} \\
    & \ge \lambda^2_{k}2^\frac{-c|\lambda_k|}{\beta  } (T+1)^\frac{c|\lambda_k|}{\beta \log_2 e} \int^{T+1}_{1} \frac{1}{ t}   \frac{1}{t^\frac{c|\lambda_k|}{\beta \log_2 e}} \mathrm{d}t \\
    & =  \lambda^2_{k}2^\frac{-c|\lambda_k|}{\beta  } (T+1)^\frac{c|\lambda_k|}{\beta \log_2 e} \int^{T+1}_{1}   \frac{1}{t^{1+\frac{c|\lambda_k|}{\beta \log_2 e}}} \mathrm{d}t \\
    & = \lambda^2_{k}2^\frac{-c|\lambda_k|}{\beta  } (T+1)^\frac{c|\lambda_k|}{\beta \log_2 e} \frac{\beta \log_2 e}{c|\lambda_k|} \lp 1 - (T+1)^{- \frac{\beta \log_2 e}{c|\lambda_k|}} \rp\\
    &=    \frac{|\lambda_{k}| \beta \log_2 e}{c}  2^\frac{-c|\lambda_k|}{\beta  }  \lp (T+1)^\frac{c|\lambda_k|}{\beta \log_2 e} -1 \rp,
    \label{eq:nc_3_lb} \numberthis
\end{align*}
where \eqref{eq:exp_trick} holds since $1+x \ge 2^x$ for all $x\in (0,1]$.
Finally, equations \eqref{eq:nc_1_lb}, \eqref{eq:nc_2_lb} and \eqref{eq:nc_3_lb} give \begin{align} \nonumber
   | \gen | & \ge \frac{c}{d \beta^2 n} \sum^d_{k=1}  \frac{|\lambda_{k}| \beta \log_2 e}{c}  2^\frac{-c|\lambda_k|}{\beta  }  \lp (T+1)^\frac{c|\lambda_k|}{\beta \log_2 e} -1 \rp \\ \nonumber
   & \geq   \frac{ \log_2 e \min_k \{ |\lambda_{k}|\}  }{ n \beta }    2^{-c}  \lp (T+1)^\frac{c\min_k \{ |\lambda_{k}|\} }{\beta \log_2 e} -1 \rp.
\end{align}  Similarly, for a smaller set of values for $c$, we may use the inequality $1+x \geq \zeta^x$ for some $\zeta\in [2,e]$ instead of the inequality $1+x \ge 2^x$ in \eqref{eq:exp_trick} and we identically find that for all $0 \le c \le 1$ such that $1+c \ge \zeta^c$, \begin{align}
   | \gen | 
   & \geq   \frac{ \log_\zeta e \min_k \{ |\lambda_{k}|\}  }{ n \beta }    \zeta^{-c}  \lp (T+1)^\frac{c\min_k \{ |\lambda_{k}|\} }{\beta \log_\zeta e} -1 \rp.
\end{align} For every choice of 
$c$, we may take the maximum allowable value of $\zeta$, namely $\zeta= (1+c)^{1/c}$. Then, together with the choice $\Lambda = -\beta \mathbf{I}$, we obtain
\begin{align*}
   | \gen | 
   & \geq   \frac{ \log_{(1+c)^{1/c}} e }{ (1+c) n  }      \lp (T+1)^\frac{c }{ \log_{(1+c)^{1/c}} e} -1 \rp\\
   &=   \frac{ c }{ (1+c)\log(1+c) n  }      \lp (T+1)^{\log (c+1)} -1 \rp\\
    &\ge   \frac{ 1 }{ 2\log(2) n  }      \lp (T+1)^{\log (c+1)} -1 \rp\\
    &\ge   \frac{ 1 }{ 2 n  }      \lp (T+1)^{\log (c+1)} -1 \rp.\numberthis
\end{align*}
The proof is now complete.\qedwhite 
\section{Strongly-Convex Loss}
Herein, we prove the generalization error bounds for strongly-convex losses. We start by proving the upper bound and then we continue with the lower bound.
\subsection{Strongly-Convex Loss Upper Bound}
The next lemma gives the contraction property of the growth recursion for strongly-convex losses.
\begin{lemma}\label{lemma:contraction_strongly}
Let the loss function be $\gamma$-strongly-convex ($\gamma >0$) and $\beta$-smooth for all $z\in\mc{Z}$. If the step size satisfies the inequality $\eta_t \leq 2/(\beta + \gamma)$, then for any $t\leq T+1$ it is true that
\begin{align*}
    &\bigg\Vert w_{t} - w^{(i)}_{t}-\eta_t 
    \bigg( \frac{1}{m}\sum_{z\in J_t}\nabla f(w_{t} , z) - \frac{1}{m}\sum_{z\in J_t}\nabla f( w^{(i)}_{t} , z) \bigg)  \bigg\Vert_2 \leq \lp 1-  \frac{\eta_t \gamma}{2}  \rp\Vert w_{t} - w^{(i)}_{t} \Vert_2 .
\end{align*}
\end{lemma}
\paragraph{Proof.}
The function $\frac{1}{m} \sum_{z\in J_t} f (\cdot, z)$ is also $\beta$-smooth and $\gamma$ strongly-convex for all $z\in\mc{Z}$ (uniformly) and the strong convexity gives\begin{align*}
    & \bigg\langle\frac{1}{m}\sum_{z\in J_t}\nabla f(w_{t} , z) - \frac{1}{m}\sum_{z\in J_t}\nabla f( w^{(i)}_{t} , z),
    w_{t} -  w^{(i)}_{t} \bigg\rangle
    \\& \geq\frac{\beta \gamma }{\beta+\gamma}  \Vert w_{t} - w^{(i)}_{t}  \Vert^2_2 + \frac{1}{(\beta+\gamma) } \Vert \frac{1}{m}\sum_{z\in J_t}\nabla f(w_{t} , z) - \frac{1}{m}\sum_{z\in J_t}\nabla f( w^{(i)}_{t} , z) \Vert^2_2. \numberthis \label{eq:cocoersive_strong_convex_loo}
\end{align*} We expand the squared norm as follows 
\begin{align*}
    &\bigg\Vert w_{t} - w^{(i)}_{t}-\eta_t 
    \bigg( \frac{1}{m}\sum_{z\in J_t}\nabla f(w_{t} , z) - \frac{1}{m}\sum_{z\in J_t}\nabla f( w^{(i)}_{t} , z) \bigg)
    \bigg\Vert^2_2\\
    &= \Vert w_{t} - w^{(i)}_{t} \Vert^2_2 - 2\eta_t   \inp{\frac{1}{m}\sum_{z\in J_t}\nabla f(w_{t} , z) - \frac{1}{m}\sum_{z\in J_t}\nabla f( w^{(i)}_{t} , z)}{w_{t} -  w^{(i)}_{t}}  \\
    &\qquad + \eta^2_t\Vert \frac{1}{m}\sum_{z\in J_t}\nabla f(w_{t} , z) - \frac{1}{m}\sum_{z\in J_t}\nabla f( w^{(i)}_{t} , z)  \Vert^2_2 \\
   &\leq  \Vert w_{t} - w^{(i)}_{t} \Vert^2_2  + \eta^2_t\Vert \frac{1}{m}\sum_{z\in J_t}\nabla f(w_{t} , z) - \frac{1}{m}\sum_{z\in J_t}\nabla f( w^{(i)}_{t} , z) \Vert^2_2  \\
    &\qquad - 2\eta_t 
    \bigg( \frac{\beta \gamma }{\beta+\gamma}  \Vert w_{t} - w^{(i)}_{t}  \Vert^2_2 + \frac{1}{(\beta+\gamma) } \Vert\frac{1}{m}\sum_{z\in J_t}\nabla f(w_{t} , z) - \frac{1}{m}\sum_{z\in J_t}\nabla f( w^{(i)}_{t} , z) \Vert^2_2 \bigg)
    \numberthis \label{eq:apply_cocersive_strong_loo}
    \\
    &= \lp 1 - 2\eta_t \frac{\beta \gamma }{\beta+\gamma} \rp \Vert w_{t} - w^{(i)}_{t} \Vert^2_2 \\& \qquad+ \eta_t\lp \eta_t - \frac{2}{\beta+\gamma} \rp  \Vert \frac{1}{m}\sum_{z\in J_t}\nabla f(w_{t} , z) - \frac{1}{m}\sum_{z\in J_t}\nabla f( w^{(i)}_{t} , z) \Vert^2_2   \\
    &\leq \lp 1 - 2\eta_t \frac{\beta \gamma }{\beta+\gamma} \rp \Vert w_{t} - w^{(i)}_{t} \Vert^2_2 . \label{eq:contraction1_loo} \numberthis
\end{align*} 
We apply the inequality \eqref{eq:cocoersive_strong_convex_loo} to derive \eqref{eq:apply_cocersive_strong_loo}. The inequality \eqref{eq:contraction1_loo} holds since $\eta_t\leq 2/(\beta+\gamma)$ and $\beta \geq \gamma $. Also \begin{align}
    2\eta_t \frac{\beta \gamma }{\beta+\gamma} &\geq 2\eta_t \frac{\beta \gamma}{2\beta} =  \eta_t \gamma . \label{eq:simplifies_ratio_loo}
\end{align} We derive the bound of the lemma through \eqref{eq:contraction1_loo}, \eqref{eq:simplifies_ratio_loo} and $\sqrt{1-x}\leq 1-x/2 $ for $x\in [0,1]$.\qedwhite

\subsection{Proof of Lemma \ref{lemma:Mini-Batch sc growth} (Growth Recursion---Strongly-Convex Loss)}\label{eq:proof_growth_strongly}
\begin{align*}
    G_{\!J_t}(w) \triangleq   w - \eta_t \frac{1}{m} \sum_{z \in J_t} \nabla f(w ,z ) ,\,\,\,
    G_{\!J^{(i)}_t}(w) \triangleq w - \eta_t \frac{1}{m} \sum_{z \in J^{(i)}_t} \nabla f(w ,z )   \numberthis
\end{align*}
Under the same batch selection $J_t \equiv J^{(i)}_t$ we find the first part of the statement as
\begin{align*}
    \Vert G_{\!J_t}(w_{t}) - G_{\!J_t}(w^{(i)}_t) \Vert  
    & \leq \bigg\Vert   w_t - \eta_t \frac{1}{m} \sum_{z \in J_t} \nabla f(w ,z )|_{w=w_t}     -   w^{(i)}_t + \eta_t \frac{1}{m} \sum_{z \in J_t} \nabla f(w ,z )|_{w=w^{(i)}_t}   \bigg\Vert \\
    &\leq (1 - \frac{\eta_t \gamma}{2} )\Vert w_t - w^{(i)}_t \Vert  ,\numberthis
\end{align*} as Lemma \ref{lemma:contraction_strongly} suggests. Similarly, we find the bound for the case $J_t \neq J^{(i)}_t $ as
\begin{align*}
    \Vert G_{\!J_t}(w_t) - G_{\!J^{(i)}_t}(w^{(i)}_t) \Vert 
     & = \Vert G_{\!J_t}(w_t) -  G_{\!J_t}(w^{(i)}_t) +  G_{\!J_t}(w^{(i)}_t) - G_{\!J^{(i)}_t}(w^{(i)}_t) \Vert 
    \\ & \leq \Vert G_{\!J_t}(w_t) -  G_{\!J_t}(w^{(i)}_t)\Vert + \Vert G_{\!J_t}(w^{(i)}_t) - G_{\!J^{(i)}_t}(w^{(i)}_t) \Vert\\
    &\leq \Vert G_{\!J_t}(w_t) -  G_{\!J_t}(w^{(i)}_t)\Vert + \frac{\eta_t}{m}\lp \Vert \nabla f(w^{(i)}_t , z_i) \Vert + \Vert \nabla f(w^{(i)}_t , z'_i) \Vert \rp \\
    &\leq (1 - \frac{\eta_t \gamma}{2} )\Vert w_t - w^{(i)}_t \Vert +\frac{\eta_t}{m}\lp \Vert \nabla f(w^{(i)}_t , z_i) \Vert + \Vert \nabla f(w^{(i)}_t , z'_i) \Vert \rp .
\end{align*} 
For both cases we have 
\begin{align*}
    &\Vert w_{t+1}- w^{(i)}_{t+1} \Vert \\
    & \leq (1 - \frac{\eta_t \gamma}{2} )\Vert w_t - w^{(i)}_t \Vert + \frac{1}{m} \lp \Vert \nabla f(w^{(i)}_t , z_i) \Vert + \Vert \nabla f(w^{(i)}_t , z'_i) \Vert \rp \eta_t \mathds{1}_{J_t \neq J^{(i)}_t }
    \\ &\leq  (1 - \frac{\eta_t \gamma}{2} )\Vert w_t - w^{(i)}_t \Vert + \frac{1}{m} \lp \sup_{t, S^{(i)}, z_i}\Vert \nabla f(w^{(i)}_t , z_i) \Vert + \sup_{t, S^{(i)}}\Vert \nabla f(w^{(i)}_t , z'_i) \Vert \rp \eta_t \mathds{1}_{J_t \neq J^{(i)}_t } \\
    &\leq  (1 - \frac{\eta_t \gamma}{2} )\Vert w_t - w^{(i)}_t \Vert + \frac{2}{m} \lp \sup_{t, S^{(i)}, z_i}\Vert \nabla f(w^{(i)}_t , z_i) \Vert  \rp \eta_t \mathds{1}_{J_t \neq J^{(i)}_t }\\
    & =  (1 - \frac{\eta_t \gamma}{2} )\Vert w_t - w^{(i)}_t \Vert + \frac{2}{m} \lp \sup_{t, S^{(i)}, z}\Vert \nabla f(w^{(i)}_t , z) \Vert  \rp \eta_t \mathds{1}_{J_t \neq J^{(i)}_t } \\
    & = (1 - \frac{\eta_t \gamma}{2} )\Vert w_t - w^{(i)}_t \Vert + \frac{2}{m} \lp \sup_{t, S^{(1)}, z}\Vert \nabla f(w^{(1)}_t , z) \Vert  \rp \eta_t \mathds{1}_{J_t \neq J^{(i)}_t } \\
    & = (1 - \frac{\eta_t \gamma}{2} )\Vert w_t - w^{(i)}_t \Vert + \frac{2}{m} \lp \sup_{t, S, z'}\Vert \nabla f(w_t , z') \Vert  \rp \eta_t \mathds{1}_{J_t \neq J^{(i)}_t }\\
    &\leq (1 - \frac{\eta_t \gamma}{2} )\Vert w_t - w^{(i)}_t \Vert + \frac{2\tilde{L}}{m} \eta_t \mathds{1}_{J_t \neq J^{(i)}_t }.\numberthis
\end{align*}
The last inequality  completes the proof. \qedwhite 
\subsection{Proof Of Theorem \ref{Thm:Lower_bound} (Lower Bound---Strongly-Convex Loss)}\label{proof_strongly_lower}
Set the initial point $w_1 = 0$ (independent of the data-set) and select the loss function as $f(w,\mathbf{z})=  ( w-\mathbf{z})^\T  \Lambda  ( w-\mathbf{z}) /2$, $w\in\mbb{R}^d$, $\Lambda \triangleq \text{diag}(\beta,\gamma,\ldots ,\gamma)$, $\mathbf{z}_i =(z^1_i, z^2_i ,\ldots ,z^d_i )\in\{ -L/\gamma \sqrt{d} ,+L/\gamma\sqrt{d} \}^{d} $, $i\in \{ 1, 2,\ldots , n \}$, $\P (z_i^j = L/\gamma\sqrt{d})= 1/2 $ and $z^\ell_k \perp z^{\ell '}_{k'} $ for all distinct pairs $(\ell,k)$, $(\ell' , k')$. Then the SGD updates with step size $\eta_t = c/(\beta+\gamma) $ for $c\leq 1$ are \begin{align}
    w_{t+1} &=w_t - \eta_t \Lambda\lp w_t-\frac{1}{m}\sum_{\mathbf{z}\in J_t } \mathbf{z}  \rp = \lp 1- \frac{c}{\beta+\gamma} \Lambda \rp w_t +  \frac{c\Lambda}{(\beta+\gamma)m}\sum_{\mathbf{z}\in J_t } \mathbf{z} . 
\end{align} 
For any initial point (independent of the data-set), the output of the algorithm is
\begin{align}
    A_{\mc{R}}(S)&\equiv w_{T+1} \nonumber\\&=  
    \frac{c}{(\beta+\gamma)m}  \sum^{T}_{t=1} \lp \prod^{T}_{j=t+1} \lp 1- \frac{c}{\beta+\gamma} \Lambda \rp \rp \Lambda \sum_{\mathbf{z}\in J_t } \mathbf{z}   \nonumber \\& =  \frac{c}{(\beta+\gamma)m}  \sum^{T}_{t=1}  \lp 1- \frac{c}{\beta+\gamma} \Lambda \rp^{T-t} \Lambda \sum_{\mathbf{z}\in J_t } \mathbf{z}.
\end{align} 
Then the norm of the gradient for any $\bar{t}\leq T$ is bounded as \begin{align*}
    \Vert \nabla f ( w_{\bar{t}+1} , \mathbf{z}) \Vert &= \bigg\Vert   \frac{c\Lambda}{(\beta+\gamma)m}  \sum^{\bar{t}}_{t=1}  \lp 1- \frac{c}{\beta+\gamma} \Lambda \rp^{\bar{t}-t} \Lambda \sum_{\mathbf{z}\in J_t } \mathbf{z}  - \Lambda \mathbf{z} \bigg\Vert \\
    & \leq \bigg\Vert \frac{c\Lambda}{(\beta+\gamma)m}  \sum^{\bar{t}}_{t=1}  \lp 1- \frac{c}{\beta+\gamma} \Lambda \rp^{\bar{t}-t} \Lambda \sum_{\mathbf{z}\in J_t }   \mathbf{z}  \bigg\Vert + \Vert  \Lambda \mathbf{z} \Vert\\
     & \leq \frac{\sqrt{\sum^d_{k=1}  \lambda^2_k   \lp 1-\lp 1- \frac{c\lambda_k}{\beta+\gamma}  \rp^{\bar{t}}\rp^2  }}{\gamma\sqrt{d}} L  + \frac{\sqrt{\beta^2 +(d-1)\gamma^2}}{\gamma\sqrt{d}} L \\
     & \leq \frac{\sqrt{\sum^d_{k=1}  \lambda^2_k     }}{\gamma\sqrt{d}} L  + \frac{\sqrt{\beta^2 +(d-1)\gamma^2}}{\gamma\sqrt{d}} L  \\
     & = 2 \frac{\sqrt{\beta^2 +(d-1)\gamma^2}}{\gamma\sqrt{d}} L  \\
     & \leq 4 L,  \label{eq:4L1} \numberthis
\end{align*} for any $d $ such that
$$\beta^2 + (d-1)\gamma^2 < 4 \gamma^2 d 
\implies 
\beta^2 - \gamma^2  < 3 \gamma^2 d
\implies d > \frac{\beta^2 -\gamma^2}{3\gamma^2}.  $$
%
%
Similarly, for any $\tilde{c}\in(0,1)$ 
\begin{align*}
    &\Vert \nabla f ( \tilde{c} A_{\mc{R}}(S) +(1-\tilde{c})A_{\mc{R}}(S^{(i)}) , \mathbf{z}) \Vert \\&= \bigg\Vert   \frac{\tilde{c}c\Lambda}{(\beta+\gamma)m}  \sum^{T}_{t=1}  \lp 1- \frac{c}{\beta+\gamma} \Lambda \rp^{T-t} \Lambda \sum_{\mathbf{z}\in J_t } \mathbf{z} +\frac{(1-\tilde{c})c\Lambda}{(\beta+\gamma)m}  \sum^{T}_{t=1}  \lp 1- \frac{c}{\beta+\gamma} \Lambda \rp^{T-t} \Lambda \sum_{\mathbf{z}\in J^{(i)}_t } \mathbf{z}  - \lambda \mathbf{z} \bigg\Vert \\
    & \leq \tilde{c}\bigg\Vert \frac{c\Lambda}{(\beta+\gamma)m}  \sum^{T}_{t=1}  \lp 1- \frac{c}{\beta+\gamma} \Lambda \rp^{T-t} \Lambda \sum_{\mathbf{z}\in J_t }   \mathbf{z} \bigg\Vert 
    \\&\quad\quad\quad+(1-\tilde{c})\bigg\Vert \frac{c\Lambda}{(\beta+\gamma)m}  \sum^{T}_{t=1}  \lp 1- \frac{c}{\beta+\gamma} \Lambda \rp^{T-t} \Lambda \sum_{\mathbf{z}\in J^{(i)}_t }   \mathbf{z} \bigg\Vert + \Vert  \Lambda \mathbf{z}\Vert\\
     & \leq \frac{\sqrt{\sum^d_{k=1}  \lambda^2_k     }}{\gamma\sqrt{d}} L + \frac{\sqrt{\beta^2 +(d-1)\gamma^2}}{\gamma\sqrt{d}} L \\
     & = 2 \frac{\sqrt{\beta^2 +(d-1)\gamma^2}}{\gamma\sqrt{d}} L \\
     & \leq 4 L. \label{eq:4L2}
\end{align*} 
Additionally, \begin{align*}
    \gen & =\E_{S,z,\mc{R}} [f(A_{\mc{R}}  (S); \mathbf{z}) -\frac{1}{n}\sum^n_{i=1} f(A_{\mc{R}}  (S); \mathbf{z}_i )]  \\
    &=   \E_{S,S',\mc{R}} \left[ \frac{1}{n}\sum^n_{i=1} f(A_{\mc{R}}  (S); \mathbf{z}'_i)] -\frac{1}{n}\sum^n_{i=1} f(A_{\mc{R}}  (S); \mathbf{z}_i ) \right] \\
    &=  \frac{1}{2} \E_{S,S',\mc{R}} \left[ \frac{1}{n}\sum^n_{i=1}  (  A_{\mc{R}} (S)-\mathbf{z}'_i)^\T  \Lambda  (  A_{\mc{R}} (S)-\mathbf{z}'_i) -\frac{1}{n}\sum^n_{i=1} (  A_{\mc{R}} (S)-\mathbf{z}_i)^\T  \Lambda  (  A_{\mc{R}} (S)-\mathbf{z}_i) \right] \\
    & = \frac{1}{2} \E_{S,S',\mc{R}} \Bigg[  \frac{1}{n}\sum^n_{i=1} \big( A^\T_{\mc{R}} (S) \Lambda A_{\mc{R}} (S) - 2A^\T_{\mc{R}} (S) \Lambda \mathbf{z}'_i  + (\mathbf{z}'_i)^\T \Lambda \mathbf{z}'_i \\&\qquad\qquad \qquad\qquad- A^\T_{\mc{R}} (S) \Lambda A_{\mc{R}} (S)  +  2 A^\T_{\mc{R}} (S) \Lambda  \mathbf{z}_i  - (\mathbf{z}_i)^\T \Lambda \mathbf{z}_i \big) \Bigg]\\
    & =     \frac{1}{2n}\sum^n_{i=1} \big(  - 2\E_{S,\mc{R}}[A^\T_{\mc{R}} (S) ] \Lambda \underbrace{\E_{S'}[\mathbf{z}'_i]}_{=0}  + \underbrace{\E_{S'}[(\mathbf{z}'_i)^\T \Lambda \mathbf{z}'_i ]  - \E_{S}[(\mathbf{z}_i)^\T \Lambda \mathbf{z}_i]}_{=0} +  \E_{S,\mc{R}} [2 A^\T_{\mc{R}} (S) \Lambda \mathbf{z}_i ]   \big) \\
    & =  \frac{1}{2n}\sum^n_{i=1}    \E_{S,\mc{R}} [2 A^\T_{\mc{R}} (S) \Lambda \mathbf{z}_i ]   \\
    & =  \frac{c}{(\beta+\gamma)nm}\sum^n_{i=1}    \E_{S,\mc{R}} \Bigg[  \mathbf{z}^T_i \Lambda   \sum^{T}_{t=1}  \lp 1- \frac{c}{\beta+\gamma} \Lambda \rp^{T-t} \Lambda    \sum_{\mathbf{z}\in J_t } \mathbf{z} \Bigg].
\end{align*}
Define
\begin{align}
    M^t\triangleq \Lambda     \lp 1- \frac{c}{\beta+\gamma} \Lambda \rp^{T-t} \Lambda.
\end{align}
If $\mathbf{z}_i \in J_t$ then $\E_{S}[ z^k_i \sum_{\mathbf{z}\in J_t } z^k ] =L^2 /\gamma^2 d$, otherwise $\E_{S}[ z^k_i \sum_{\mathbf{z}\in J_t } z^k ] =0 $, thus $\E_{S}[ z^k_i \sum_{\mathbf{z}\in J_t } z^k ]= \mathds{1}_{\mathbf{z}_i \in J_t} L^2/\gamma^2 d  $ for any (randomized) selection rule, and $\sum^n_{i=1} \mathds{1}_{\mathbf{z}_i \in J_t} = m$ for any $A_{\mc{R}}\in \setalgos$ (Definition \ref{alg:gen_SGD}) with batch-size $m\equiv |J_t|$ for all $t$. Therefore, we obtain 
\begin{align*}
    \frac{c}{(\beta+\gamma)nm}\E_\mc{R} \left[ \sum^n_{i=1}  \sum^d_{k=1}\sum^T_{t=1} M^t_{kk} \E_{S} [  z^k_i  \sum_{\mathbf{z}\in J_t } z^k ] ]     \right] & = \frac{c}{\gamma^2(\beta+\gamma)nm}\E_\mc{R} \left[ \sum^n_{i=1}  \sum^d_{k=1} \sum^T_{t=1} M^t_{kk} \mathds{1}_{\mathbf{z}_i \in J_t} L^2/d  \right] \\
     & = \frac{c L^2}{\gamma^2d(\beta+\gamma)nm}\E_\mc{R} \left[  \sum^d_{k=1} \sum^T_{t=1} M^t_{kk} \sum^n_{i=1} \mathds{1}_{\mathbf{z}_i \in J_t}   \right] \\
     & = \frac{c L^2}{\gamma^2d(\beta+\gamma)n}\E_\mc{R} \left[  \sum^d_{k=1} \sum^T_{t=1} M^t_{kk}    \right]\\
     & = \frac{c L^2}{\gamma^2d(\beta+\gamma)n} \sum^d_{k=1}\sum^T_{t=1} M^t_{kk}.   
\end{align*} Then we find $M_{kk}$ as \begin{align*}
    \sum^T_{t=1} M^t_{kk}&= \lambda^2_k  \sum^{T}_{t=1} \prod^{T}_{j=t+1} \lp 1- \frac{c\lambda_k}{\beta+\gamma}  \rp^j \\
    &= \lambda^2_k  \sum^{T}_{t=1}  \lp 1- \frac{c\lambda_k}{\beta+\gamma}  \rp^{T-t}
    \\
    &= \lambda^2_k \lp 1- \frac{c\lambda_k}{\beta+\gamma}  \rp^{T}  \sum^{T}_{t=1}  \lp 1- \frac{c\lambda_k}{\beta+\gamma}  \rp^{-t}\\
    &=  \lambda^2_k \lp 1- \frac{c\lambda_k}{\beta+\gamma}  \rp^{T}  \frac{1- \lp 1- \frac{c\lambda_k}{\beta+\gamma} \rp^{-T}}{1-(1- \frac{c\lambda_k}{\beta+\gamma})^{-1}} \lp 1- \frac{c\lambda_k}{\beta+\gamma} \rp^{-1}\\
    &=  \lambda^2_k   \frac{\lp 1- \frac{c\lambda_k}{\beta+\gamma}  \rp^{T}- 1}{  1- \frac{c\lambda_k}{\beta+\gamma} -1} \\
    &=   \frac{\lambda_k }{c} (\beta+\gamma)  \lp 1-\lp 1- \frac{c\lambda_k}{\beta+\gamma}  \rp^{T}\rp.\numberthis
\end{align*}
 Thus
 \begin{align*}
     \gen &= \frac{L^2}{\gamma^2d(\beta+\gamma)n}  \sum^d_{k=1} \sum^T_{t=1} M^t_{kk}   \\
     &= \frac{cL^2}{\gamma^2d(\beta+\gamma)n}  \sum^d_{k=1} \frac{\lambda_k}{c} (\beta+\gamma)  \lp 1-\lp 1- \frac{c\lambda_k}{\beta+\gamma}  \rp^{T}\rp \\
     &= \frac{L^2}{\gamma^2d n}  \sum^d_{k=1} \lambda_k   \lp 1-\lp 1- \frac{c\lambda_k}{\beta+\gamma}  \rp^{T}\rp \\
     &= \frac{L^2}{\gamma^2d n}  \sum^d_{k=1} \lambda_k     -  \frac{L^2}{\gamma^2 d n}  \sum^d_{k=1}  \lambda_k  \lp 1- \frac{c\lambda_k}{\beta+\gamma}  \rp^{T} \\
     &\geq \frac{L^2}{\gamma n}      -  \frac{L^2}{\gamma^2 d n}  \sum^d_{k=1} \lambda_k  \lp 1- \frac{c\lambda_k}{\beta+\gamma}  \rp^{T} \\
     &= \frac{L^2}{\gamma n}      -  \frac{L^2 (\beta +\gamma)}{c\gamma^2 d n}  \sum^d_{k=1} \frac{c\lambda_k }{\beta+\gamma}  \lp 1- \frac{c\lambda_k}{\beta+\gamma}  \rp^{T} \\
     &\geq  \frac{L^2}{\gamma n}      -  \frac{L^2 (\beta +\gamma)}{c\gamma^2  n}  \frac{1 }{T+1}  \lp 1- \frac{1}{T+1}  \rp^{T} \\
     &= \frac{L^2}{\gamma n}      -  \frac{L^2 (\beta +\gamma)}{c\gamma^2  n}  \frac{1 }{T+1}  \lp \frac{T }{T+1}  \rp^{T} \\
     &\geq \frac{L^2}{\gamma n}      -  \frac{L^2 (\beta +\gamma)}{c\gamma^2  n}  \frac{1 }{T+1}  \\
      &= \frac{L^2}{\gamma n}    \lp 1  -  \frac{ (\beta +\gamma)}{c\gamma}  \frac{1 }{T+1}\rp\\
      &= \frac{L^2}{\gamma n}    \lp 1  -  \frac{ 1}{\eta_t \gamma}  \frac{1 }{T+1}\rp\label{eq:gen_last_sc}\numberthis
 \end{align*}Note that the last expression is non-trivial for all $\eta_t > 1/\gamma (T+1)$. For $2/\gamma (T+1)\leq \eta_t =c/(\beta+\gamma)$, let $\Tilde{L}\equiv 4 L$ ($L$ is a free parameter in the example above) and it is true that $\Lf\leq \Tilde{L} $ due to the inequalities  \eqref{eq:4L1} and \eqref{eq:4L2} (and $\Lf$ is defined in \eqref{eq:Lf}). Thus for 
 $d \geq (\beta^2 -\gamma^2) /3\gamma^2$ 
 the inequality \eqref{eq:gen_last_sc} holds for all $A_{\mc{R}}\in \setalgos$ and the smallest value of $\eta_t$ gives\begin{align}
   \inf_{A_{\mc{R}}\in \setalgos} \sup_{\substack{(f,\mc{D})\in \mc{SC}^{\tilde{L}}_{\beta,\gamma} } }  |\gen (f , \mc{D},A_{\mc{R}} ) |\geq \frac{\tilde{L}^2}{32\gamma n},
\end{align}
as claimed.\qedwhite

\section{Inherent Limitations of Uniform Stability
}\label{failure}
Herein, we demonstrate that the proof techniques in~\cite{hardt2016train} produce vacuous generalization bounds for incremental gradient methods~\cite[Algorithm 3]{pmlr-v125-safran20a},~\cite{bertsekas1999nonlinear,bertsekas2015convex,Nedić2001,doi:10.1137/17M1147846}, for general Lipschitz and smooth (strongly) convex losses.
\subsection{Incremental Gradient Method: Convex Loss}

Let the batch size be $m=1$ (since $m=1$ has been considered in~\cite{hardt2016train}).
Then the update at time $t$ has the form 
$$    w_{t+1} = w_t -\eta_t \nabla f(w_t , z_{J_t} ),$$ and $J_t = (t-1) \mod n +1 \in [1,T]$ ($T$ the total number of iterations) is the sample that the algorithm selects at time. Consider the neighboring sequences $S= (z_1 , z_2 , \ldots , z_{i^*} ,\ldots ,z_n)$ and $S^{\prime} \equiv S^{(i^*)} = (z_1 , z_2 , \ldots, z_{i^*}^{\prime} ,\ldots ,z_n)$ (where $i^*$ is   $1\leq i^* \leq n$).  Define $z_{J_t}^{\prime}$ as any element in $S'$. 
Note that the mappings $G_t(\cdot)$ and $G_t(\cdot)^{\prime}$ are 
$$  G_t (w)  =   w -\eta_t \nabla f(w , z_{J_t} ),\quad \text{for } z_{J_t}\in S  $$
 $$  G_t^{\prime} (w)  =   w -\eta_t \nabla f(w , z_{J_t}^{\prime} ), \quad \text{for } z_{J_t}^{\prime}  \in S' .$$ 
At this point we will consider cases for clarity.

\paragraph{Case 1.} $T<n$. Since the step-size sequence is a non-increasing the mappings have to be different at time $t=1$, as we prove below. In fact, the method in~\cite{hardt2016train} for convex loss gives
\begin{align*} \sup_{S ,S' , z}    \Vert  w_T - w_T^{\prime} \Vert &= \sup_{S ,S^{(i^*)} , z} \Vert G_{T-1}(w_{T-1} ) - G_{T-1}^{\prime} (w_{T-1}^{\prime} )  \Vert \\
    & =  \sup_{S ,S^{(i^*)} , z} \Vert G_{T-1}(w_{T-1} ) - G_{T-1}^{\prime} (w_{T-1}^{\prime} ) \Vert   
    \\
    & \leq  \sup_{S ,S^{(i^*)} , z} \Vert w_{T-1}  - w_{T-1}^{\prime} \Vert  \\
    & =  \sup_{S ,S^{(i^*)} , z} \Vert G_{T-2}(w_{T-2} ) - G_{T-2}^{\prime} (w_{T-2}^{\prime} ) \Vert  \\
    & =  \sup_{S ,S^{(i^*)} , z} \Vert G_{T-2}(w_{T-2} ) - G_{T-2}(w_{T-2}^{\prime} ) \Vert  \\
    & \leq  \sup_{S ,S^{(i^*)} , z} \Vert w_{n-2}  - w_{n-2}^{\prime} \Vert   \\
    & \quad \vdots \\
    & =  \sup_{S ,S^{(i^*)} , z} \Vert G_{1}(w_{1} ) - G_{1}^{\prime}  (w_{1}^{\prime} ) \Vert \quad \text{ because of the } \sup \text{ we have here }  G_{1}(\cdot)\neq G_{1}^{\prime} (\cdot) \\
    & \leq \Vert w_{1}  - w_{1}^{\prime} \Vert +2L  \eta_{1} \\ 
    &  \leq \Vert w_{0}  - w_{0}^{\prime}  \Vert + 2L \eta_{1} \\
    & =2L \eta_{1} .
\end{align*} 
Then the generalization error in~\cite{hardt2016train} is bounded as
\begin{align}
     \sup_{S ,S^{\prime} , z} \mathbb{E}_A [ f(A(S),z)- f(A(S^{\prime}),z)] \leq L \sup_{S ,S^{\prime} , z}  \Vert  w_n - w_n^{\prime} \Vert \leq  2L^2 \eta_{1}.
 \end{align} 
On the other hand we have established that (Theorem \ref{thm:OnAVER_C}) $$ |\epsilon_{\text{gen}}|\leq \frac{2L^2}{n} \sum^T_{t=1} \eta_t,$$
which is the optimal generalization bound for the incremental gradient algorithm,
while the technique in~\cite{hardt2016train} gives an evidently vacuous result.
\paragraph{Case 2.} $T=K\times n$, where $K$ is the total number of epochs. 
Let the step-size be $\eta_t$. Then similarly to the Case 1 above by induction we get
\begin{align}
    \sup_{S,S',z} \Vert w_T - w'_T \Vert \leq 2L \sum^{K-1}_{k=0} \eta_{kn+1}.
\end{align} As a consequence the method in~\cite{hardt2016train} gives \begin{align}
    \sup_{S ,S^{\prime} , z} \mathbb{E}_A [ f(A(S),z)- f(A(S^{\prime}),z)] \leq L \sup_{S ,S^{\prime} , z}  \Vert  w_n - w_n^{\prime} \Vert \leq  2L^2 \sum^{K-1}_{k=0} \eta_{kn+1} .
\end{align} For instance, for a step-size choice $\eta_t = \beta^{-1} /((t-1)\mod n +1) $ (which we restart at the end of each epoch), the last expression gives \begin{align}
    \sup_{S ,S^{\prime} , z} \mathbb{E}_A [ f(A(S),z)- f(A(S^{\prime}),z)] \leq L \sup_{S ,S^{\prime} , z}  \Vert  w_n - w_n^{\prime} \Vert \leq  2L^2 K \eta_{1}.
    \end{align}
This is of order $\mc{O}(1)$, because $K$ is an integer. On the other hand we have established the optimal bound 
\begin{align}
    |\epsilon_{\text{gen}}|\leq \frac{2L^2}{n} \sum^T_{t=1} \eta_t = \frac{2L^2}{n} \sum^{Kn}_{t=1} \eta_t = \frac{2KL^2}{n} \sum^{n}_{t=1} \frac{1}{\beta t}\leq \frac{2L^2 K(\log (n)+1)}{\beta n},
\end{align}  
which is evidently of order $\mathcal{O}(\log(n)/n)$.

\paragraph{Case 3.} $T=K\times n$, where $K$ is the total number of epochs. Let the step-size be fixed $\eta_t = \eta $.\\
Then our lower bound (Theorem \ref{Thm:Lower_bound_convex}) gives
\begin{align}
    |\eps_{\text{gen}}|\geq \frac{L^2}{2n} \sum^{Kn}_{t=1} = K \frac{\eta L^2}{2} =\Omega (1),
\end{align}and eventually the upper bounds in~\cite{hardt2016train} and in our work are of order $\mc{O}(1)$. The problem is hopeless with such a choice of the step-size and for such a large number of iterations for all batch schedules. Note that this impossibility result is new.

\subsection{Incremental Gradient Method: $\gamma$-Strongly-Convex Loss}
In the case of strongly-convex losss (the gradient mappings are now $1-\eta \gamma$ expansive \cite{hardt2016train}), we choose a fixed step size $\eta_t = \eta$, and $T=K\times n$ (for $K$ number of epochs). Then, for the method in~\cite{hardt2016train}, we consider the following two cases. 
\paragraph{Case 1.} One epoch, $T= n$. We can expand as
\begin{align*}
    \Vert  w_n - w_n^{\prime} \Vert &= \Vert G_{n-1}(w_{n-1} ) - G_{n-1}^{\prime} (w_{n-1}^{\prime} )  \Vert \\
    & = \Vert G_{n-1}(w_{n-1} ) - G_{n-1}(w_{n-1}^{\prime} ) \Vert \\
& \leq (1-\eta \gamma)\Vert w_{n-1}  - w_{n-1}^{\prime} \Vert \\
    & = (1-\eta \gamma)\Vert G_{n-2}(w_{n-2} ) - G_{n-2}^{\prime} (w_{n-2}^{\prime} ) \Vert \\
    & = (1-\eta \gamma)\Vert G_{n-2}(w_{n-2} ) - G_{n-2}(w_{n-2}^{\prime} ) \Vert \\
    & \leq (1-\eta \gamma)^2 \Vert w_{n-2}  - w_{n-2}^{\prime} \Vert\\
    & \quad \vdots \\
    & \leq (1-\eta \gamma)^{n-i^*-2}\Vert G_{i^{*}+1}(w_{i^{*}+1} ) - G_{i^{*}+1}^{\prime} (w_{i^{*}+1}^{\prime} ) \Vert \\
    & = (1-\eta \gamma)^{n-i^*-2}\Vert G_{i^{*}+1}(w_{i^{*}+1} ) - G_{i^{*}+1}(w_{i^{*}+1}^{\prime} ) \Vert \\
    & \leq (1-\eta \gamma)^{n-i^* -1}\Vert w_{i^{*}+1}  - w_{i^{*}+1}^{\prime}  \Vert \\
& = (1-\eta \gamma)^{n-i^* -1}\Vert G_{i^{*}}(w_{i^{*}} ) - G_{i^{*}}^{\prime} (w_{i^{*} }^{\prime} ) \Vert \quad \text{ here } G_{i^{*}}(\cdot)\neq G_{i^{*}}^{\prime} (\cdot) \\
& \leq (1-\eta \gamma)^{n-i^*}\Vert w_{i^{*}}  - w_{i^{*}}^{\prime} \Vert + 2L\eta (1-\eta \gamma)^{n-i^*-1} \\
    & = (1-\eta \gamma)^{n-i^*} \Vert G_{i^{*}-1}(w_{i^{*}-1} ) - G_{i^{*}-1}^{\prime} (w_{i^{*}-1}^{\prime} ) \Vert + 2L\eta (1-\eta \gamma)^{n-i^*-1} \\
    & = (1-\eta \gamma)^{n-i^*} \Vert G_{i^{*}-1}(w_{i^{*}-1} ) - G_{i^{*}-1}(w_{i^{*}-1}^{\prime} ) \Vert + 2L\eta (1-\eta \gamma)^{n-i^*-1}\\
    & \leq (1-\eta \gamma)^{n-i^* +1}\Vert w_{i^{*}-1}  - w'_{i^{*}-1}  \Vert + (1-\eta \gamma)^{i^*} 2L\eta (1-\eta \gamma)^{n-i^*-1}\\
    & \quad \vdots \\
    &  \leq (1-\eta \gamma)^{n}   \Vert w\_{0}  - w\_{0}^{\prime}  \Vert +  2L \eta (1-\eta \gamma)^{n-i^*-1} \\
    & =  2L\eta (1-\eta \gamma)^{n-i^*-1}.
        \end{align*}
Then the approach in~\cite{hardt2016train} gives the following vacuous bound
\begin{align*}
    \sup_{S ,S^{\prime} , z} \mathbb{E}_A [ f(A(S),z)- f(A(S^{\prime}),z)] \leq L \sup_{S ,S^{\prime} , z}  \Vert  w_n - w_n^{\prime} \Vert \leq 2L^2 \sup_{i^*} (1-\eta \gamma)^{n-i^*-1} = 2L^2 \eta (1-\eta \gamma)^{-1}.
 \end{align*} 
The above bound is of order $\mc{O}(1)$. In contrast, we prove the optimal rate in Theorem \ref{thm:OnAVER_SC}.
\paragraph{Case 2.} Multiple epochs, $T= K\times n$.
By induction on Case 1 and the approach in~\cite{hardt2016train}, we obtain \begin{align*}
    \sup_{S ,S^{\prime} , z} \mathbb{E}_A [ f(A(S),z)- f(A(S^{\prime}),z)] \leq L \sup_{S ,S^{\prime} , z}  \Vert  w_n - w_n^{\prime} \Vert \leq 2L^2 \sup_{i^*} \sum^K_{k=1} (1-\eta \gamma)^{kn-i^*-1} = \mc{O} (1) .
 \end{align*} In contrast, we prove the correct rate in Theorem \ref{thm:OnAVER_SC}. Our bounds are of the order $\mc{O}(1/n)$ which is the optimal generalization rate in this case.

\end{document}

%% file: macros.tex
\newcommand\numberthis{\addtocounter{equation}{1}\tag{\theequation}}

\newcommand{\lp}{\left(}
\newcommand{\rp}{\right)}

\newcommand{\mc}{\mathcal}

\newcommand{\mbb}{\mathbb}

\newcommand{\gen}{\eps_{\mathrm{gen}}}

\usepackage[normalem]{ulem}
\newcommand{\cn}{\textcolor{red}{[\raisebox{-0.2ex}{\tiny\shortstack{citation\\[-1ex]needed}}]}}

\newcommand{\kn}[1]{\textcolor{blue}{\textbf{KN:} #1}}
\newcommand{\dk}[1]{\textcolor{red}{\textbf{DK:} #1}}

\newcommand\redout{\bgroup\markoverwith{\textcolor{red}{\rule[.5ex]{2pt}{0.4pt}}}\ULon}

\renewcommand{\P}{\mbb{P}}
\newcommand{\E}{\mbb{E}}

\newcommand{\T}{\mathrm{T}}

\newcommand{\eps}{\epsilon}

\newcommand{\Vast}{\bBigg@{5}}
\newcommand{\setalgos}{\mc{A}(\eta_t,T)}
\newcommand{\Lf}{\Tilde{L}_{\eta_t,T}(f)}

\newcommand{\mds}{\mathds}

\newcommand{\epath}{\eps_{\mathrm{path}}}